\documentclass[lettersize,journal]{IEEEtran}
\usepackage[T1]{fontenc}
\usepackage{graphicx}
\usepackage{amsmath,amssymb}
\usepackage{bm}
\usepackage{booktabs}
\usepackage{multirow}
\usepackage[table,dvipsnames]{xcolor}
\usepackage{pifont}
\usepackage{enumitem}
\usepackage[caption=false,font=normalsize,labelfont=sf,textfont=sf]{subfig}
\usepackage{stfloats}
\usepackage{float}
\usepackage{array}
\usepackage{makecell}
\usepackage{tabularx}
\usepackage{colortbl}
\usepackage{arydshln}
\usepackage{cite}
\usepackage{placeins}
\usepackage{hyperref}
\usepackage{etoolbox}
\AtBeginEnvironment{tabular}{\footnotesize}
\AtBeginEnvironment{tabularx}{\footnotesize}

\definecolor{tableheader}{HTML}{D6E4F0}
\definecolor{taskheader}{HTML}{FFF4E6}
\definecolor{oursrow}{HTML}{E2EFDA}
\definecolor{bestcol}{HTML}{1F4E79}
\definecolor{checkgreen}{HTML}{2E7D32}
\definecolor{avgcol}{HTML}{FFF0A0}
\renewcommand{\arraystretch}{0.88}
\newcommand{\cmark}{{\color{checkgreen}\ding{51}}}
\newcommand{\xmark}{{\color{gray!50}\ding{55}}}
\newcommand{\best}[1]{\textbf{\color{red!70!black}#1}}
\newcommand{\second}[1]{\underline{\color{bestcol}#1}}

\newcommand{\modelname}{Earth-OneVision}
\newcommand{\datasetname}{MMRS-OneVision}
\newcommand{\acsa}{ACSA}

\title{\modelname: Extending Remote Sensing Multimodal Large Language Models to More Sensor Modalities and Tasks}
\author{Miaoxin Cai,~\IEEEmembership{Graduate Student Member,~IEEE,} Guanqun Wang,~\IEEEmembership{Member,~IEEE,} Wei Zhang,~\IEEEmembership{Member,~IEEE,} Guangyao Zhou, Yin~Zhuang\textsuperscript{\dag},~\IEEEmembership{Member,~IEEE,} Tong Zhang, Hao Wang,~\IEEEmembership{Member,~IEEE,} He Chen,~\IEEEmembership{Member,~IEEE,} and Jun~Li,~\IEEEmembership{Fellow,~IEEE}
\thanks{This work was supported in part by the Ye Qisun Science Foundation of the National Natural Science Foundation of China, under Grant U2341202, in part by the General Program of National Natural Science Foundation of China under Grant 62371048, and in part by the Fund of the National Key Laboratory of Space-Based Intelligent Information Processing under Grant tj012504. (\textsuperscript{\dag}Corresponding author: Yin Zhuang.)}
\thanks{Miaoxin Cai, Guanqun Wang, He Chen, and Yin Zhuang are with the National Key Laboratory of Science and Technology on Space-Born Intelligent Information Processing (SBIIP), Beijing Institute of Technology, Beijing 100081, China (e-mail: yzhuang@bit.edu.cn).}
\thanks{Guangyao Zhou is with the Aerospace Information Research Institute, Chinese Academy of Sciences, Beijing 100094, China, and also with the Key Laboratory of Technology in Geo-Spatial Information Processing and Application System, Chinese Academy of Sciences, Beijing 100190, China.}
\thanks{Wei Zhang is with the Advanced Research Institute of Multidisciplinary Sciences, Beijing Institute of Technology, Beijing 100081, China, and also with the School of Mechatronical Engineering, Beijing Institute of Technology, Beijing 100081, China.}
\thanks{Hao Wang is with the School of Earth and Space Sciences, Peking University, Beijing 100871, China.}
\thanks{Tong Zhang is with the School of Electronics, Peking University, Beijing 100871, China.}
\thanks{Jun Li is with the School of Computer Science and Hubei Key Laboratory of Intelligent Geo-Information Processing, China University of Geosciences, Wuhan, 430078, China.}}

\begin{document}

\markboth{}{}

\maketitle

\begin{abstract}

RS-MLLMs enable natural-language understanding and spatial reasoning over earth observation imagery. However, existing models support only a narrow range of sensor types and tasks, yielding a fragmented view of the earth and leaving cross-modal geoscientific knowledge largely unexploited. This work presents \modelname{}, a 2B RS-MLLM that unifies six sensor modalities (i.e., optical, SAR, infrared, multispectral, temporal, and video) and cross-sensor fusion across 9 task categories within a single autoregressive framework. Three dedicated mechanisms address three bottlenecks. Full-Granularity Vision-Language Alignment (FGVLA) aligns multi-level visual features with the multi-dimensional language space. Spatial-Linguistic Isomorphic Serialization (SLIS) unifies heterogeneous spatial outputs as autoregressive tokens. Progressive Cross-Modality Adaptation (PCMA) decomposes the compound domain gap into sequential stages, tackling the viewpoint and imaging physics gaps in turn. To support joint training, \datasetname{} is constructed with $\sim$34M QA pairs spanning all six sensor modalities and cross-sensor fusion across 9 task categories, substantially exceeding existing RS multimodal instruction datasets. With only 2B parameters, \modelname{} achieves competitive or state-of-the-art results across extensive benchmarks, consistently matching or outperforming 4B-72B RS-MLLMs. It achieves 87.52\% P@0.5 on the OPT-RSVG testset for optical visual grounding and 80.68\% on the SAR VQA benchmark SARLANG-Bench, exceeding 7B models by over 7\%. It further achieves 75.74\% recall on the BigEarthNet-MS testset for multispectral classification, and 81.94\% MCQ accuracy on EarthMind-Bench for cross-modality reasoning.

\end{abstract}

\begin{IEEEkeywords}
Remote sensing, multimodal large language model, unified model, cross-task, cross sensor-modality.
\end{IEEEkeywords}

\section{Introduction}
\begin{figure}[!t]
    \centering
    \includegraphics[width=\columnwidth]{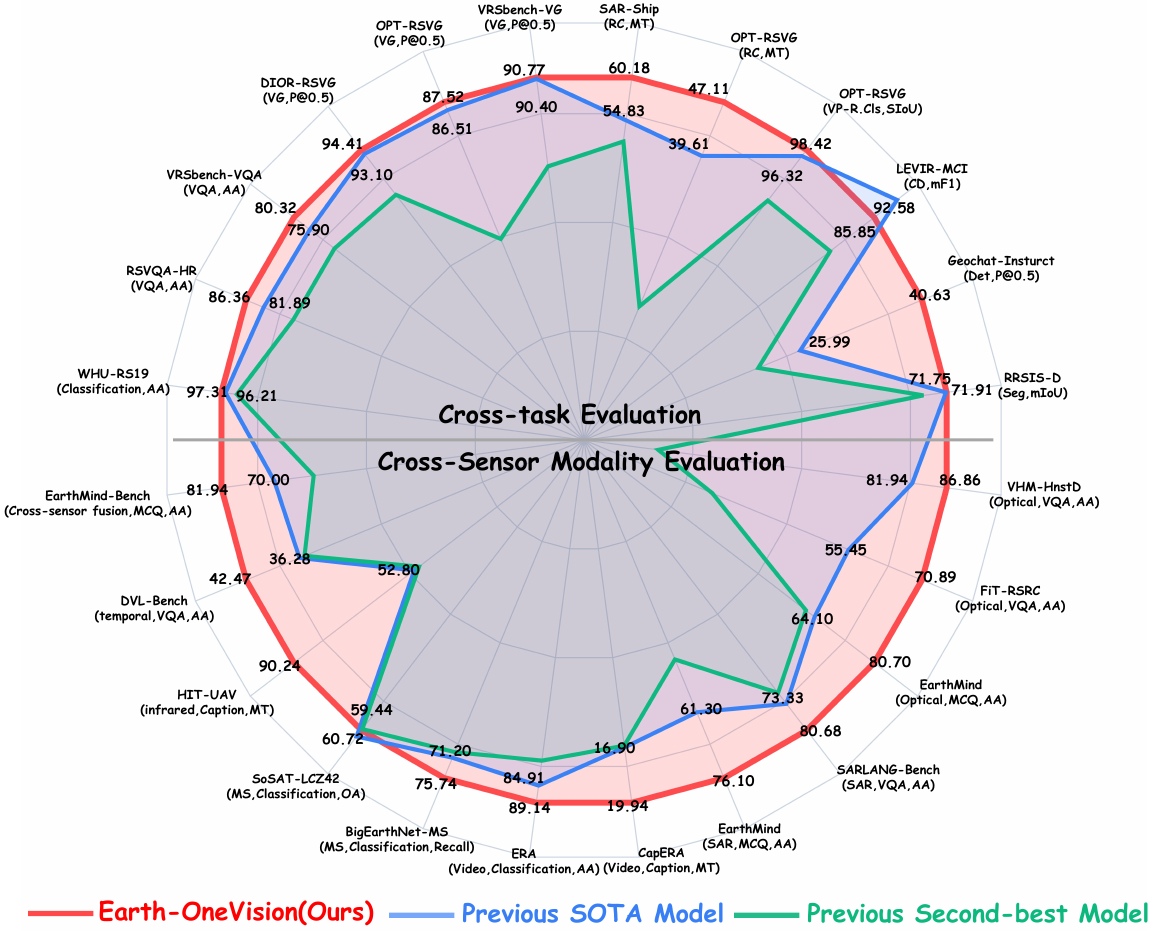}
    \caption{Performance of \modelname{} vs.\ state-of-the-art across 24 cross-task and cross sensor-modality benchmarks. \modelname{} (red circle) consistently matches or surpasses previous SOTA model (blue) and previous second-best model (green) across all benchmarks.}
    \label{fig:radar}
    \end{figure}

\IEEEPARstart{R}{emote} sensing (RS) earth observation acquires surface information through diverse sensors and has been widely applied in disaster response, urban planning, and environmental monitoring~\cite{al2024integrating}. Modern RS platforms produce multi-sensor, multi-modal, and multi-temporal data. Each sensor type encodes a distinct physical signal: optical captures reflected light, SAR measures microwave backscatter, infrared detects thermal emission, and multispectral resolves fine spectral differences. Meanwhile, interpretation demands have grown increasingly diverse, encompassing scene classification, object detection, visual grounding, change detection, semantic segmentation, visual question answering (VQA), and more. However, existing methods typically address only 1-3 sensor types and 3-5 task categories, fragmenting knowledge into isolated silos and leaving cross-modal and cross-task synergies largely unexploited.

\begin{figure*}[t]
\centering
\includegraphics[width=\textwidth]{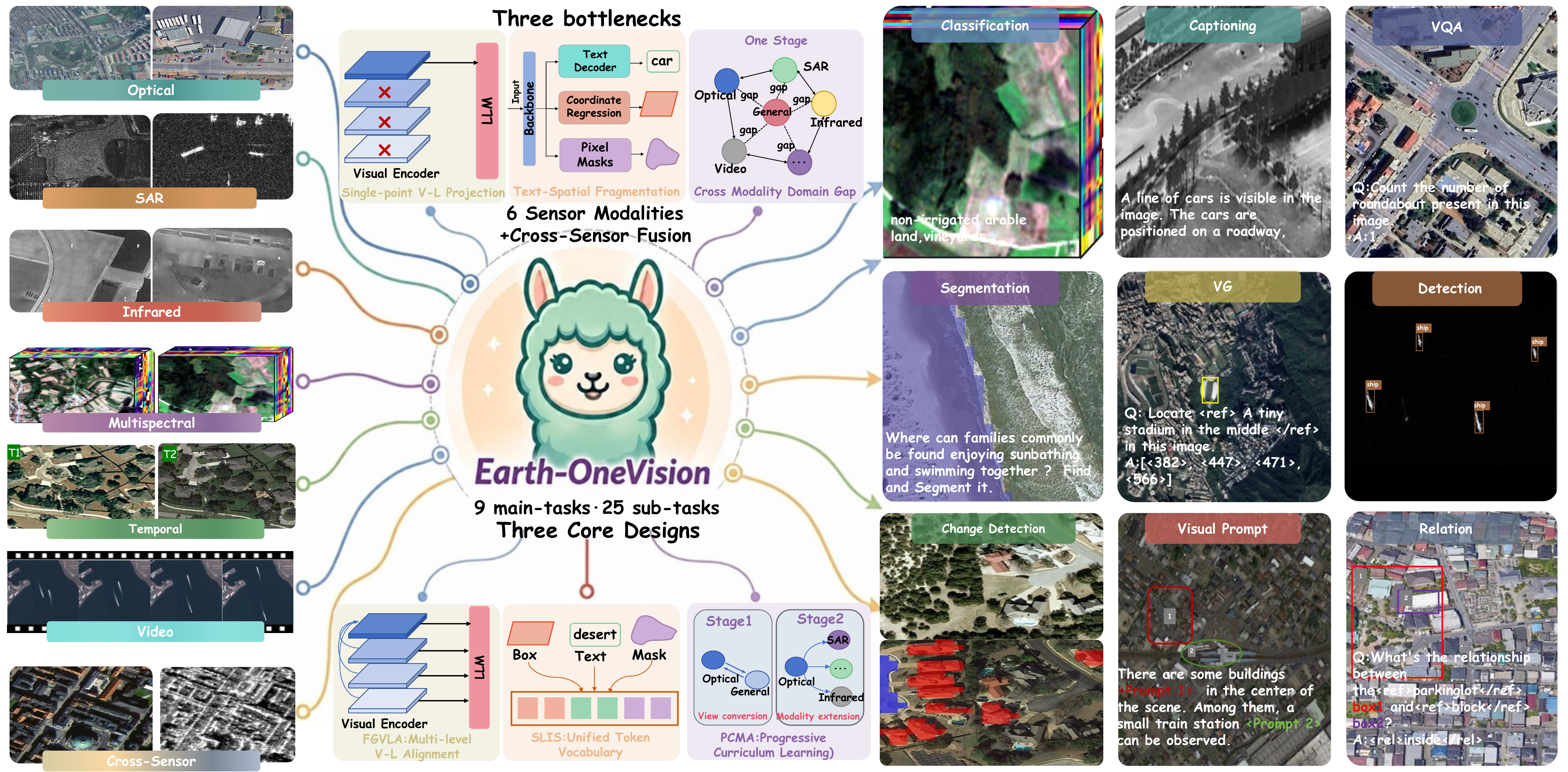}
\caption{\modelname{} overview: six sensor modalities and cross-sensor fusion unified as input, three designs addressing three bottlenecks (i.e., FGVLA for vision-language alignment, SLIS for spatial-linguistic fragmentation, PCMA for cross-modality transfer), supporting 9 task categories.}
\label{fig:teaser}
\end{figure*}

Multimodal large language models (MLLMs) have achieved notable progress: BLIP-2~\cite{li2023blip2}, LLaVA~\cite{liu2024visual}, Qwen-VL~\cite{bai2023qwen}, InternVL~\cite{chen2024internvl2}, LLaVA-OneVision~\cite{li2024llavaonevision}, and LISA~\cite{lai2024lisa} advance vision-language alignment, instruction tuning, and pixel-level understanding. However, RS images differ substantially from natural images in imaging geometry, viewing angle, and object characteristics, which prevents direct transfer.

To bridge this gap, RS-specific multimodal large language models (RS-MLLMs) have emerged: RSGPT~\cite{hu2023rsgpt} for captioning and VQA, GeoChat~\cite{kuckreja2023geochat} for multi-task grounded dialogue, EarthGPT~\cite{zhang2024earthgpt} for multi-modal perception, SkySenseGPT~\cite{luo2024skysensegpt} for relation reasoning, and UniGeoSeg~\cite{ni2025unigeoseg} and DVLChat~\cite{xuan2025dynamicvl} for pixel-level tasks via external decoders. EarthDial~\cite{soni2025earthdial} attempts broader sensor coverage but trains separate optical and non-optical models. TEOChat~\cite{irvin2024teochat}, EarthMarker~\cite{zhang2025earthmarker}, and EarthGPT-X~\cite{zhang2025earthgpt} extend temporal and visual prompting capabilities. Yet all these models project only final-layer features to a single LLM entry point and employ heterogeneous spatial output mechanisms with no unified generation paradigm.

These limitations reveal three bottlenecks current RS-MLLMs have yet to overcome. \textbf{Shallow vision-language alignment}: existing RS-MLLMs couple vision and language at a single point, where the final ViT layer is projected once into the LLM's input, discarding intermediate visual representations and bypassing the LLM's internal depth, leaving fine-grained spatial and semantic cues mutually inaccessible. \textbf{Fragmented task paradigms}: heterogeneous spatial outputs (i.e., horizontal boxes, oriented boxes, and segmentation masks) rely on separate decoder pathways, fracturing the shared representation space and hindering joint optimization. \textbf{Cross-modality transfer gap}: RS transfer requires crossing two compound gaps (i.e., a viewpoint gap between natural and overhead imagery, and an imaging physics gap between optical and non-optical modalities such as visible-light reflection, microwave backscatter, and thermal radiation), which makes direct transfer highly challenging.

To address these bottlenecks, this work presents \textbf{\modelname{}}, as illustrated in Fig.~\ref{fig:teaser}, an RS-MLLM built on a single autoregressive framework supporting six sensor modalities (i.e., optical, SAR, infrared, multispectral, temporal, and video) and cross-sensor fusion, across 9 task categories. Targeting the architecture, vocabulary, and training dimensions, \modelname{} introduces three dedicated mechanisms while preserving the foundation model's pretrained natural-scene knowledge. (i) \textbf{Full-Granularity Vision-Language Alignment (FGVLA)} bridges the multi-level visual hierarchy and the multi-dimensional language space, enabling comprehensive vision-language interaction beyond single-layer feature projection. (ii) \textbf{Spatial-Linguistic Isomorphic Serialization (SLIS)} serializes heterogeneous spatial formats (i.e., horizontal boxes, oriented boxes, and segmentation masks) into language-like token sequences, unifying all spatial and linguistic tasks under a single generative paradigm. (iii) \textbf{Progressive Cross-Modality Adaptation (PCMA)} bridges two compound transfer gaps, viewpoint and imaging physics, through staged adaptation, progressively extending from natural-scene knowledge to optical RS and then to non-optical modalities. To support joint multi-modality and multi-task training, \textbf{\datasetname{}} is constructed with over 34M QA pairs spanning six sensor modalities and cross-sensor fusion across 9 task categories.

Evaluations across 24 RS benchmarks show that \modelname{} consistently rivals or surpasses 4B-72B counterparts, as illustrated in Fig.~\ref{fig:radar}, demonstrating the synergistic benefits of unified cross sensor-modality multi-task learning and matching or exceeding 4B-72B models despite its 2B scale.

The main contributions are:
\begin{itemize}
    \item \textbf{Broad-spectrum multi-modality and multi-task RS-MLLM.} \modelname{} is the first single-architecture RS-MLLM that unifies six sensor modalities and cross-sensor fusion across 9 task categories within a single 2B-parameter autoregressive framework.
    \item \textbf{Broad-spectrum multi-modality architecture.} Three dedicated mechanisms (i.e., FGVLA, SLIS, and PCMA) resolve the shallow vision-language alignment, fragmented task paradigms, and cross-modality transfer gap bottlenecks within a single autoregressive framework.
    \item \textbf{Large-scale RS multimodal instruction dataset.} \datasetname{} is constructed with approximately 34M QA pairs spanning six sensor modalities and cross-sensor fusion across 9 task categories, exceeding prior RS multimodal instruction datasets in scale, modality breadth, and task coverage, enabling unified multi-task training.
    \item \textbf{Competitive performance with fewer parameters.} \modelname{} consistently matches or surpasses 4B-72B RS-MLLMs across cross-task and cross sensor-modality benchmarks, validating the effectiveness of the proposed unified framework.
\end{itemize}

\begin{table*}[t]
    \centering
    \caption{Sensor-modality coverage and task capability of existing RS-MLLMs. EarthDial uses separate Opt and Non-Opt models. Opt=Optical, SAR=Synthetic Aperture Radar, IR=Infrared, MS=Multispectral, Temp=Temporal, Vid=Video, Fus=Cross-sensor Fusion, Cls=Classification, Det=Detection, VG=Visual Grounding, VQA=Visual Question Answering, Cap=Captioning, Seg=Segmentation, CD=Change Detection, VP=Visual Prompting, RR=Relation Reasoning.}
    \label{tab:rs_mllm_comparison}
    \setlength{\tabcolsep}{10pt}
    \renewcommand{\arraystretch}{1.2}
    \resizebox{\textwidth}{!}{
    \tiny
    \begin{tabular}{l|ccccccc|ccccccccc}
    \toprule
    \multirow{2.4}{*}{\textbf{MLLM}} & \multicolumn{7}{c|}{\cellcolor{tableheader}\textbf{Sensor Input}} & \multicolumn{9}{c}{\cellcolor{taskheader}\textbf{Task Type}} \\
     & \cellcolor{tableheader}Opt & \cellcolor{tableheader}SAR & \cellcolor{tableheader}IR & \cellcolor{tableheader}MS & \cellcolor{tableheader}Temp & \cellcolor{tableheader}Vid & \cellcolor{tableheader}Fus & \cellcolor{taskheader}Cls & \cellcolor{taskheader}Det & \cellcolor{taskheader}VG & \cellcolor{taskheader}VQA & \cellcolor{taskheader}Cap & \cellcolor{taskheader}Seg & \cellcolor{taskheader}CD & \cellcolor{taskheader}VP & \cellcolor{taskheader}RR \\
    \midrule
    \mbox{GeoChat~\cite{kuckreja2023geochat}} & \cmark & \xmark & \xmark & \xmark & \xmark & \xmark & \xmark & \cmark & \xmark & \cmark & \cmark & \cmark & \xmark & \xmark & \xmark & \xmark \\
    \mbox{SkySenseGPT~\cite{luo2024skysensegpt}} & \cmark & \xmark & \xmark & \xmark & \xmark & \xmark & \xmark & \xmark & \cmark & \cmark & \cmark & \cmark & \xmark & \xmark & \xmark & \cmark \\
    \mbox{EarthGPT~\cite{zhang2024earthgpt}} & \cmark & \cmark & \cmark & \xmark & \xmark & \xmark & \xmark & \cmark & \cmark & \cmark & \cmark & \cmark & \xmark & \xmark & \xmark & \xmark \\
    \mbox{EarthDial (Opt)~\cite{soni2025earthdial}} & \cmark & \xmark & \xmark & \xmark & \xmark & \xmark & \xmark & \cmark & \cmark & \cmark & \cmark & \cmark & \xmark & \xmark & \xmark & \xmark \\
    \mbox{EarthDial (Non-Opt)~\cite{soni2025earthdial}} & \xmark & \cmark & \cmark & \cmark & \cmark & \xmark & \xmark & \cmark & \cmark & \cmark & \cmark & \cmark & \xmark & \xmark & \xmark & \xmark \\
    \mbox{TEOChat~\cite{irvin2024teochat}} & \cmark & \xmark & \xmark & \xmark & \cmark & \xmark & \xmark & \cmark & \cmark & \cmark & \cmark & \cmark & \xmark & \cmark & \xmark & \xmark \\
    \mbox{EarthGPT-X~\cite{zhang2025earthgpt}} & \cmark & \cmark & \cmark & \xmark & \xmark & \xmark & \xmark & \cmark & \xmark & \xmark & \cmark & \cmark & \xmark & \xmark & \cmark & \xmark \\
    \mbox{DVLChat~\cite{xuan2025dynamicvl}} & \cmark & \xmark & \xmark & \xmark & \cmark & \xmark & \xmark & \xmark & \xmark & \xmark & \cmark & \cmark & \cmark & \cmark & \xmark & \xmark \\
    \mbox{UniGeoSeg~\cite{ni2025unigeoseg}} & \cmark & \xmark & \xmark & \xmark & \xmark & \xmark & \xmark & \xmark & \xmark & \xmark & \xmark & \xmark & \cmark & \xmark & \xmark & \xmark \\
    \mbox{RSThinker~\cite{liu2025rsthinker}} & \cmark & \xmark & \xmark & \xmark & \xmark & \xmark & \xmark & \cmark & \cmark & \cmark & \cmark & \cmark & \xmark & \xmark & \xmark & \xmark \\
    \midrule
    \rowcolor{oursrow}
    \textbf{\modelname{} (Ours)} & \cmark & \cmark & \cmark & \cmark & \cmark & \cmark & \cmark & \cmark & \cmark & \cmark & \cmark & \cmark & \cmark & \cmark & \cmark & \cmark \\
    \bottomrule
    \end{tabular}
    }
    \end{table*}
\section{Related Work}

\subsection{Multimodal Large Language Models}

Representative LLMs such as GPT~\cite{openai2023gpt4} and LLaMA~\cite{touvron2023llama} establish strong language reasoning through large-scale pretraining. MLLMs extend these to vision via visual encoders and alignment modules. BLIP-2~\cite{li2023blip2} bridges visual and language spaces via Q-Former. LLaVA~\cite{liu2024visual} introduces visual instruction tuning as the mainstream paradigm. Qwen2.5-VL~\cite{bai2025qwen25vl} and InternVL2 and InternVL2.5~\cite{chen2024internvl2} advance dynamic resolution and progressive training. LLaVA-OneVision~\cite{li2024llavaonevision} unifies single-image, multi-image, and video training. Despite strong general performance, these models are trained on natural scene data and transfer poorly to RS: overhead viewing alters object morphologies, scales span two orders of magnitude~\cite{xia2018dota}, and SAR, infrared, and multispectral sensors yield data markedly different from RGB imagery~\cite{zhu2017deep,ma2019deep,zhu2021deep}, making direct transfer to RS interpretation difficult~\cite{kuckreja2023geochat}.

\subsection{Remote Sensing Multimodal Large Language Models}

Traditional RS interpretation relied on task-specific models for detection~\cite{xie2021oriented}, VQA~\cite{zhang2023multi}, and captioning~\cite{hu2023rsgpt}. Unified RS-MLLMs have since emerged: GeoChat~\cite{kuckreja2023geochat} builds the first multi-task optical RS dialogue model. SkySenseGPT~\cite{luo2024skysensegpt} adds relation reasoning. SkyEyeGPT~\cite{zhan2024skyeyegpt} supports multi-granularity captioning. These remain confined to optical imagery. EarthGPT~\cite{zhang2024earthgpt} extends to SAR and infrared but omits pixel-level tasks. EarthDial~\cite{soni2025earthdial} broadens sensor coverage via separate optical and non-optical models rather than a unified architecture. TEOChat~\cite{irvin2024teochat} pioneers temporal earth observation. DVLChat~\cite{xuan2025dynamicvl} and UniGeoSeg~\cite{ni2025unigeoseg} address pixel-level tasks but lack broader multi-task support. EarthMarker~\cite{zhang2025earthmarker} and EarthGPT-X~\cite{zhang2025earthgpt} introduce visual prompting but cover limited modalities.

Table~\ref{tab:rs_mllm_comparison} reveals that no single model simultaneously achieves broad modality coverage, scene-to-pixel task support across 9 task categories, and a unified spatial output paradigm. On modality, the broadest unified single-architecture coverage reaches only three sensor types. Multispectral, video, and cross-sensor fusion remain absent. On tasks, VQA and captioning are near-universal, but coverage drops sharply for spatial tasks: detection, segmentation, change detection, and visual prompting are supported only by a minority. EarthMind~\cite{earthmind2025} and CROMA~\cite{fuller2023croma} show that optical-SAR joint modeling yields clear accuracy gains over single-modality baselines, suggesting a truly unified RS-MLLM could unlock substantially greater potential.

\subsection{RS Multimodal Instruction Datasets and Unified Task Modeling}

\textbf{Data.} SkyScript~\cite{wang2024skyscript} supplies large-scale image-caption pairs for pretraining. For instruction tuning, MMRS-1M~\cite{zhang2024earthgpt} provides million-scale multi-modal data. FIT-RS~\cite{luo2024skysensegpt} and GeoChat-Instruct~\cite{kuckreja2023geochat} target fine-grained understanding. RSVP~\cite{zhang2025earthmarker} and M-RSVP~\cite{zhang2025earthgpt} enrich visual prompting and multi-sensor coverage. Most datasets cover only a few modalities, with gaps in multispectral, temporal, and video annotations and limited pixel-level support.

\textbf{Task modeling.} Pix2Seq~\cite{chen2022pix2seq} and Shikra~\cite{chen2023shikra} reformulate spatial outputs as autoregressive coordinate sequences. Ferret~\cite{you2023ferret} supports arbitrary-shape referring. LISA~\cite{lai2024lisa} attaches an external segmentation decoder, breaking the autoregressive paradigm. PolyFormer~\cite{liu2023polyformer} explores polygon serialization for segmentation. ReX-Omni~\cite{gao2025rexomni} unifies detection, referring, and visual prompting via discrete coordinate tokens, yet no single scheme covers horizontal boxes, oriented boxes, and segmentation masks within one autoregressive vocabulary.

\textbf{Training strategy.} LLaVA~\cite{liu2024visual} uses two-stage align-then-finetune. InternVL2~\cite{chen2024internvl2} introduces progressive training. LLaVA-OneVision~\cite{li2024llavaonevision} designs a three-stage single-image, multi-image, and video curriculum. In RS, EarthGPT~\cite{zhang2024earthgpt} mixes modalities without staged adaptation, and EarthDial~\cite{soni2025earthdial} adopts a three-stage curriculum but without progressive cross-modality adaptation in a unified model.

In summary, RS-MLLM unification faces three intertwined challenges. Single-point feature injection leaves multi-level visual-language spaces unaligned. Heterogeneous spatial outputs rely on fragmented decoder pathways. Large cross-modality domain gaps lack systematic progressive adaptation, with no existing dataset covering broad modalities and tasks simultaneously. \modelname{} addresses all three by aligning multi-level visual features with the language model, unifying spatial outputs within a single autoregressive vocabulary, bridging cross-modality gaps through staged training, and constructing a large-scale instruction dataset spanning six sensor modalities and cross-sensor fusion across 9 task categories.

\begin{figure*}[t]
    \centering
    \includegraphics[width=\textwidth, height=0.5\textheight]{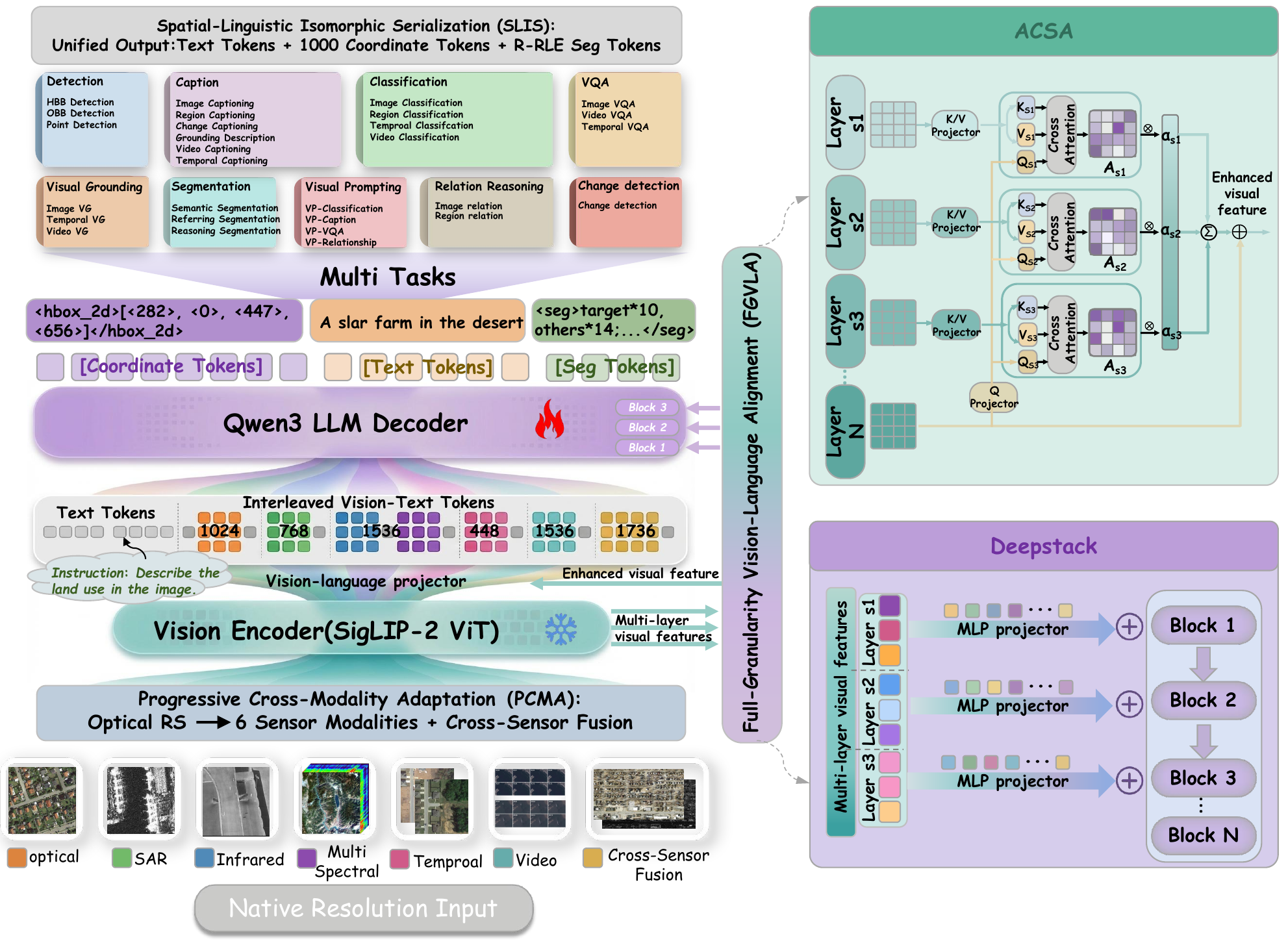}
    \caption{\modelname{} architecture: six sensor modalities and cross-sensor fusion unified as input, SigLIP-2 ViT encoder with \acsa{} aggregating multi-level features and DeepStack injecting them into early LLM layers, producing unified outputs of natural language, coordinate tokens, and R-RLE masks.}
    \label{fig:architecture}
    \end{figure*}
    
\section{Method}

Adapting a foundation MLLM to broad-spectrum RS interpretation requires overcoming three bottlenecks: single-point vision-language coupling, spatial-linguistic fragmentation, and cross-modality transfer gap. \textbf{FGVLA} routes multi-level encoder features into different LLM depths, as detailed in Section~\ref{sec:acsa}. \textbf{SLIS} reformulates horizontal boxes, oriented boxes, and masks as autoregressive token sequences, as detailed in Section~\ref{sec:spatial}. \textbf{PCMA} bridges the RS domain gap through staged training, as detailed in Section~\ref{sec:curriculum}.
\begin{figure*}[!t]
    \centering
    \includegraphics[width=\textwidth, height=0.42\textheight]{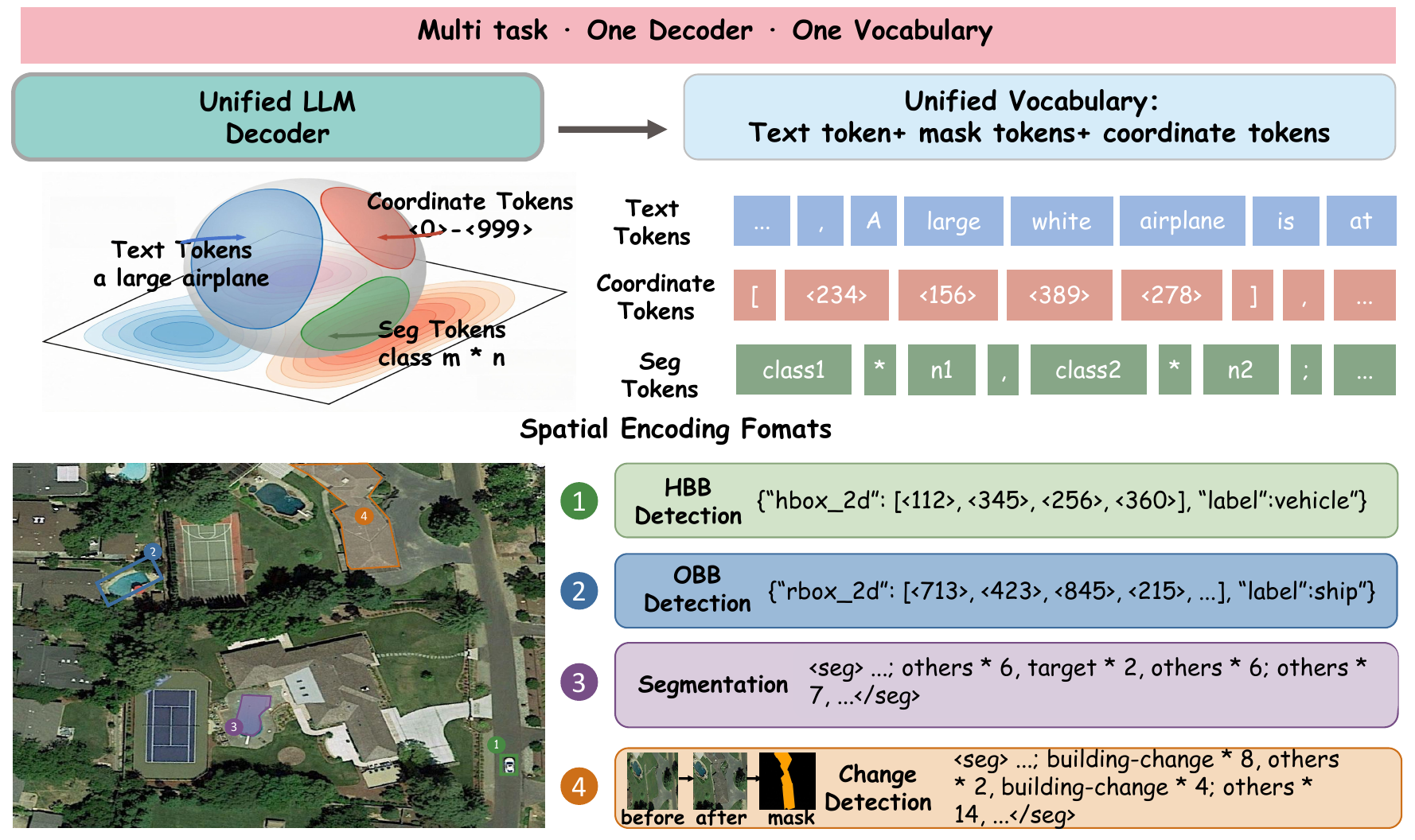}
    \caption{SLIS: text, coordinate, and seg tokens unified under one autoregressive decoder, with spatial formats including horizontal box (4 tokens), oriented box (8 tokens), and R-RLE pixel-level encoding.}
    \label{fig:spatial_encoding}
    \end{figure*}
\subsection{Model Overview}
\label{sec:overview}

\modelname{} comprises four modules: visual encoder $\mathcal{E}$, full-granularity vision-language alignment (FGVLA), vision-language projector $\mathcal{M}$, and LLM $\mathcal{G}$, as illustrated in Fig.~\ref{fig:architecture}. Given an RS image $\bm{I}$ and instruction $\bm{T}$:
\begin{equation}
    \bm{y} = \mathcal{G}\!\left(\mathcal{M}(\tilde{\bm{F}}^N),\; \{\phi_s(\bm{F}^{l_s})\}_{s=1}^{S},\; \bm{T}\right)
    \label{eq:pipeline}
\end{equation}
where $\tilde{\bm{F}}^N$ and $\{\bm{F}^{l_s}\}$ are the final and intermediate features from $\mathcal{E}(\bm{I})$ enriched by FGVLA, $\phi_s(\bm{F}^{l_s})$ are injected into early LLM layers, and $\bm{y}$ is an autoregressively generated token sequence of natural-language text, spatial coordinate tokens, seg tokens, or any mixture thereof.

The visual encoder employs SigLIP-2 (0.3B) with $N$ ViT blocks and native dynamic resolution. The vision-language projector is a two-layer MLP that compresses four adjacent visual tokens into one via $2{\times}2$ spatial merging and projects them into the LLM hidden space. The LLM is Qwen3 (2B)~\cite{yang2025qwen3}, achieving unified text-spatial outputs through vocabulary extension and token encoding, described in Section~\ref{sec:spatial}. Each input image, frame, or temporal phase is independently encoded and projected to produce visual tokens.
For multi-image inputs (i.e., multi-temporal image pairs, temporal sequences, sampled video frames, or multi-source combinations), all visual token sequences are concatenated sequentially. They are fed to the LLM together with the text instruction tokens. Training follows a progressive cross-modality adaptation curriculum, described in Section~\ref{sec:curriculum}.

\textbf{Modality input:} \modelname{} accepts six RS sensor modalities (i.e., optical, SAR, infrared, multispectral, temporal, and video) and cross-sensor fusion. Rather than introducing modality-specific encoders or dedicated fusion modules as in EarthDial~\cite{soni2025earthdial}, all modalities are unified into two encoding modes through input reorganization. \textbf{Single-image} mode handles optical directly and replicates SAR and infrared to pseudo-RGB. \textbf{Multi-image} mode independently encodes each image, groups multispectral bands into pseudo-RGB, and range-samples video frames. No modality-specific components are required.
\subsection{Full-Granularity Vision-Language Alignment}
\label{sec:acsa}

Most RS-MLLM architectures couple vision and language at a single point: the final ViT layer is projected once into the LLM's input, leaving intermediate ViT representations unused and the LLM's internal layers without direct visual access. FGVLA extends this to a \textbf{multi-depth, multi-injection design}: on the encoder side, \acsa{}, shown in the upper right of Fig.~\ref{fig:architecture}, aggregates $S{=}3$ intermediate ViT layers $\{l_1, l_2, l_3\}$ into the final representation before the projector. On the decoder side, DeepStack~\cite{meng2024deepstack}, shown in the lower right of Fig.~\ref{fig:architecture}, injects each intermediate layer directly into a corresponding early LLM layer. The result is alignment across $S{+}1$ encoder depths and $S$ LLM injection points, rather than a single projection. Both components draw from the same intermediate layers but operate at structurally independent points.

\textbf{Adaptive Cross-Scale Attention.} \acsa{} uses $\bm{F}^N$ as a query to retrieve information from each intermediate layer via cross-attention. A per-scale learnable bias $\Delta\bm{q}_s$ steers the query, and low-rank projections control parameter overhead:
\begin{equation}
    \bm{Q}_s = \bm{F}^N\bm{W}_Q + \Delta\bm{q}_s
    \label{eq:acsa_q}
\end{equation}
\begin{equation}
    \bm{K}_s = \bm{F}^{l_s}\bm{W}_{K,s}^{\mathrm{d}}\bm{W}_{K,s}^{\mathrm{u}}, \quad \bm{V}_s = \bm{F}^{l_s}\bm{W}_{V,s}^{\mathrm{d}}\bm{W}_{V,s}^{\mathrm{u}}
    \label{eq:acsa_kv}
\end{equation}
\begin{equation}
    \bm{A}_s = \operatorname{Softmax}\!\left(\frac{\bm{Q}_s\,\bm{K}_s^{\top}}{\sqrt{d_h}}\right) \bm{V}_s
    \label{eq:acsa_attn}
\end{equation}
where $s \in \{1,2,3\}$, and $\bm{W}_{K,s}^{\mathrm{d}}, \bm{W}_{V,s}^{\mathrm{d}} \in \mathbb{R}^{d \times r}$ are low-rank projections ($r \ll d$). Outputs are fused via learnable weights $\boldsymbol{\alpha}{=}\operatorname{Softmax}(\bm{w})$ and added residually:
\begin{equation}
    \tilde{\bm{F}}^N = \bm{F}^N + \operatorname{LN}(\bm{Z}\bm{W}_O), \quad \bm{Z} = \sum_{s=1}^{S} \alpha_s \bm{A}_s
    \label{eq:acsa_output}
\end{equation}
where $\bm{W}_O$ is the output projection matrix.
$\tilde{\bm{F}}^N$ replaces $\bm{F}^N$ as input to the MLP projector. DeepStack then distributes these cues across the LLM's depth.

\textbf{DeepStack Visual-Language Route.} Following Qwen3-VL~\cite{bai2025qwen3vl}, DeepStack~\cite{meng2024deepstack} equips each of $S$ intermediate ViT layers with an MLP projector $\phi_s(\cdot)$ and injects features into early LLM layers via residual addition:
\begin{equation}
    \bm{H}_s^{\text{LLM}} \leftarrow \bm{H}_s^{\text{LLM}} + \phi_s(\bm{F}^{l_s}), \quad \phi_s: \mathbb{R}^{L \times d} \rightarrow \mathbb{R}^{L/4 \times d'}
    \label{eq:deepstack}
\end{equation}
where $\bm{H}_s^{\text{LLM}}$ denotes the hidden states at the early LLM layer corresponding to the $s$-th visual level, giving the LLM direct access to different ViT depths.
\subsection{Spatial-Linguistic Isomorphic Serialization}
\label{sec:spatial}

RS spatial perception requires three structurally distinct output formats (i.e., horizontal boxes, oriented boxes, and pixel-level masks). Conventional approaches address this heterogeneity with format-specific decoders (e.g., regression heads for boxes, SAM-style mask decoders for segmentation), fragmenting the output space and precluding joint optimization under a single loss, as illustrated in Fig.~\ref{fig:spatial_encoding}.

Despite their apparent heterogeneity, all three formats admit a common reduction under appropriate discretization: spatial positions map to discrete bins, and pixel-level masks compress to run-length sequences, both yielding finite token sequences drawn from a fixed vocabulary. Generating a box, a mask, or a word then reduces to the same operation: selecting the next token under cross-entropy loss. This \emph{spatial-linguistic isomorphism} maps all spatial formats into the language token space, enabling spatial and linguistic signals to be jointly modeled within a single autoregressive sequence. The model generates mixed outputs in one pass, reasoning about \emph{where} and \emph{what} simultaneously. A unified vocabulary formalizes this:
\begin{equation}
    \mathcal{V} = \mathcal{V}_{\text{text}} \cup \mathcal{V}_{\text{coord}} \cup \mathcal{V}_{\text{seg}}
    \label{eq:slis_vocab}
\end{equation}
where $\mathcal{V}_{\text{text}}$ is the base language vocabulary, $\mathcal{V}_{\text{coord}} = \{0,1,\ldots,999\}$ consists of dedicated special tokens (i.e., not text digit strings) encoding normalized spatial positions as 1000 uniform bins, and $\mathcal{V}_{\text{seg}}$ encodes R-RLE run tokens for pixel-level masks. All three token types share the same decoder and cross-entropy loss, with no task-specific output heads.

\textbf{Coordinate tokens.} A normalized coordinate $c \in [0,1]$ is quantized to bin $q(c) = \lfloor c \times 1000 \rfloor \in \{0,\ldots,999\}$, mapping each spatial position to exactly one class in $\mathcal{V}_{\text{coord}}$. A horizontal box is serialized as four tokens $[x_1, y_1, x_2, y_2]$, and an oriented box as eight tokens $[x_1, y_1, \ldots, x_4, y_4]$. Coordinate prediction is thus a 1000-class classification under cross-entropy loss: each position maps to exactly one bin, eliminating the sequential error accumulation of character-level generation and decoupling spatial coordinates from natural-language numerals at the vocabulary level.

\textbf{R-RLE seg tokens.} A segmentation mask $M \in \mathbb{Z}^{H \times W}$ is downsampled to a $24{\times}24$ label grid $\hat{M}$ via max pooling, then row-wise run-length encoded: each row is represented as $\langle\textit{label} \times \textit{count}\rangle$ pairs, each a single entry in $\mathcal{V}_{\text{seg}}$. For each run, the model classifies the semantic category and its extent, mirroring the single-step classification logic of coordinate tokens, and reducing pixel-level prediction to region-level category classification under the same autoregressive loss. Segmentation and change detection thereby share the same vocabulary, loss, and decoder as all other tasks.

\subsection{Progressive Cross-Modality Adaptation}
\label{sec:curriculum}

Transferring general-purpose MLLMs to RS requires crossing two compound gaps: a \textbf{viewpoint gap} (i.e., natural images to optical RS, eye-level vs.\ overhead) and an \textbf{imaging physics gap} (i.e., optical RS to non-optical modalities, including visible-light reflection, microwave backscatter, and thermal radiation). Since $d(p_\text{nat}, p_\text{opt}) < d(p_\text{nat}, p_\text{all})$ and $d(p_\text{opt}, p_\text{all}) < d(p_\text{nat}, p_\text{all})$, sequential adaptation decomposes one large distribution shift into two smaller, more tractable steps. PCMA bridges the two gaps sequentially: $\mathcal{F}_1$ first closes the viewpoint gap from $p_\text{nat}$ to $p_\text{opt}$, then $\mathcal{F}_2$ closes the physics gap from $p_\text{opt}$ to $p_\text{all}$.

\textbf{Stage 1} trains jointly on natural-scene instruction data and optical RS instruction data. Optical RS shares visible-light imaging physics with natural images, differing primarily in viewpoint and object morphology, making it the closest RS modality to $p_\text{nat}$. This stage establishes RS-specific spatial understanding (e.g., overhead viewing, ground object recognition, and spatial reasoning).

\textbf{Stage 2} introduces non-optical modalities (i.e., SAR, infrared, multispectral, temporal, video, and cross-sensor fusion) and expands task coverage. Since Stage 1 has established spatial semantics, Stage 2 focuses on adapting to different imaging physics while mixing optical data to maintain the learned capabilities.

All training stages share a single autoregressive cross-entropy loss: $\mathcal{L} = -\sum_{t \in \mathcal{R}} \log P(y_t \mid y_{<t}, \bm{x})$, where $\mathcal{R}$ is the index set of response tokens.

\section{\datasetname{} Dataset}
\label{sec:dataset}

No existing RS instruction dataset supports the breadth of \modelname{}: most cover only one to three sensor modalities and lack the full spatiotemporal task spectrum. \datasetname{} fills this gap with over 34M QA pairs covering six sensor modalities and cross-sensor fusion across 9 task categories and 25 subtasks, unifying optical, SAR, infrared, multispectral, temporal, video, and cross-sensor fusion within a single dataset.

\subsection{Dataset Distribution and Characteristics}
\label{sec:dataset_overview}
\begin{figure}[!t]
    \centering
    \includegraphics[width=\columnwidth]{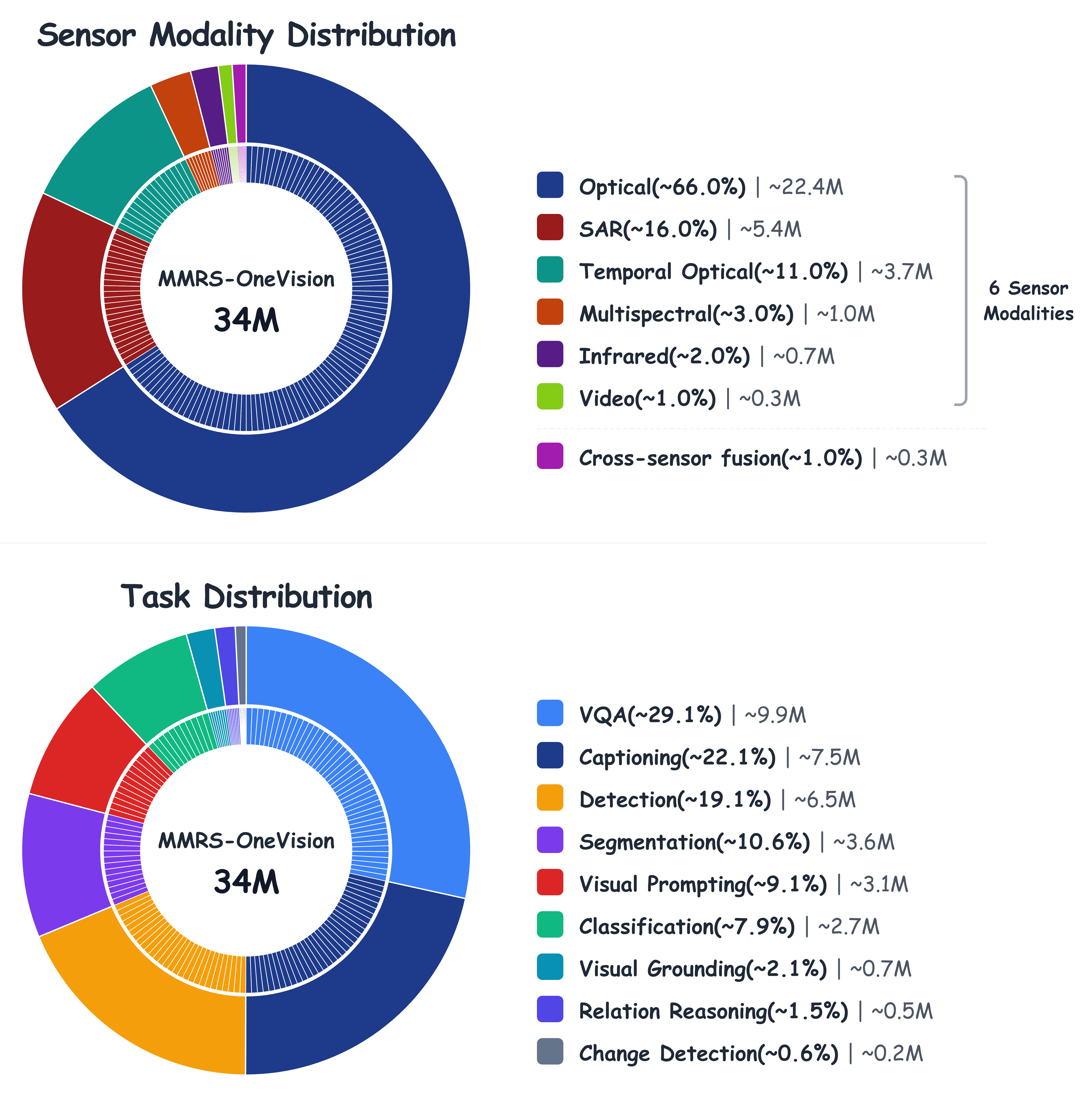}
    \caption{\datasetname{} dataset distribution. (a) Modality distribution by QA count, (b) Task distribution with inner ring showing 9 major tasks and outer ring showing 25 subtasks.}
    \label{fig:dataset_distribution}
    \end{figure}
Table~\ref{tab:dataset_detail} and Figure~\ref{fig:dataset_distribution} summarize the distribution. Optical (66\%), SAR (16\%), and temporal (11\%) dominate. Multispectral, infrared, fusion, and video serve as specialized complements. VQA (9.9M), captioning (7.5M), and detection (6.5M) account for over 70\% of training signal, with segmentation, visual prompting, classification, grounding, change detection, and relation reasoning covering the rest.

The construction process has three categories, as shown in Table~\ref{tab:dataset_detail}. \textbf{Benchmark collection (C)} directly integrates existing RS instruction datasets, including FIT-RS~\cite{luo2024skysensegpt}, SARLANG-1M~\cite{wei2026sarlang}, GeoChat~\cite{kuckreja2023geochat}, EarthMind-Instruct~\cite{earthmind2025}, VRSBench~\cite{li2024vrsbench}, EarthDial-Instruct~\cite{soni2025earthdial}, etc. \textbf{Annotation format conversion (T)} converts public RS dataset annotations into instruction data. This is the primary method. \textbf{GPT-assisted synthesis (G)} uses GPT-4o to generate richer semantic content constrained by existing annotations (Section~\ref{sec:gpt_synthesis}). A key strategy involves cross-task annotation reuse: source annotations are repurposed across tasks with different instruction templates, target outputs, and task semantics, substantially increasing the effective training volume from fixed annotation resources.

\subsection{Unified Spatial Output Format}
\label{sec:spatial_format}

Detection, grounding, segmentation, change detection, region captioning, region classification, and relation reasoning all require spatial outputs encoded as token sequences. All spatial primitives share the same vocabulary as text tokens, enabling seamless interleaving within a single autoregressive generation process.

\textbf{Detection and Grounding.}
Spatial outputs are encoded in a JSON format pairing coordinate tokens with semantic labels. Coordinates are normalized to $[0,999]$ and mapped to dedicated spatial tokens: horizontal boxes use 4 tokens, oriented boxes use 8 tokens, and points use 2 tokens. Multiple objects are listed as a JSON array. For visual grounding, the output contains the coordinates of the referred target.

\textbf{Region Captioning, Region Classification, and Relation Reasoning.}
The region of interest is specified via coordinate tags in the instruction. The model generates the corresponding textual description, category label, or relation analysis as plain text output.

\textbf{Segmentation and Change Detection (i.e., Row-wise Run-Length Encoding, R-RLE).}
Pixel-level masks are downsampled to an $h \times w$ grid (default $h{=}w{=}24$) via max pooling, then row-wise run-length encoded. The result is serialized as {\small\texttt{<seg>label *count, label *count; $\cdots$</seg>}}, where semicolons separate rows and commas separate runs. For referring segmentation, labels are \texttt{target}/\texttt{others}. For semantic segmentation and change detection, category-specific labels (e.g., \texttt{building-change}, \texttt{no-change}) are used.

\subsection{GPT-Assisted Data Synthesis}
\label{sec:gpt_synthesis}

GPT-4o synthesizes a portion of \datasetname{} QA pairs to supplement semantic content that is difficult to derive from existing annotations, adhering to one core principle: \textbf{using existing annotations as factual anchors}. Generation is always constrained by pre-existing labels, coordinates, or brief descriptions, introducing no new categories or spatial coordinates beyond the source. Four strategies cover different input types: \textbf{single-image synthesis} (i.e., image captioning and VQA from categories or brief annotations), \textbf{multi-temporal synthesis} (i.e., change captioning and temporal VQA from LEVIR-MCI, WHU-CDC, SECOND, etc.), \textbf{visual prompting synthesis} (i.e., VP-VQA across optical, SAR, infrared, and temporal modalities), and \textbf{video synthesis} (i.e., video descriptions and VQA from key frames of ERA, Sat-MTB, etc.). Synthesized data undergo two-stage quality control: rule-based filtering followed by InternVL3-72B~\cite{zhu2025internvl3} semantic review against original annotations.

\subsection{Data Construction by Type}
\label{sec:data_construction}

The following describes each data type by task. Table~\ref{tab:dataset_detail} provides the complete source dataset inventory.

\begin{table*}[!t]
\centering
\caption{Detailed source dataset inventory of \datasetname{} organized by subtask, covering six sensor modalities and cross-sensor fusion across 25 subtasks with around 34M QA pairs in total. Sensor: sensor input type covering six sensor modalities and cross-sensor fusion; O=Optical, S=SAR, I=Infrared, M=Multispectral, T=Temporal, C=Cross-sensor Fusion, V=Video. Method: \textbf{F}=format conversion, \textbf{G}=GPT synthesis, \textbf{B}=benchmark collection.}
\label{tab:dataset_detail}
\resizebox{\textwidth}{!}{
\renewcommand{\arraystretch}{1.05}
\tiny\linespread{1.15}\selectfont
\begin{tabular}{m{1.8cm}m{2.6cm}m{1.3cm}m{0.8cm}m{20cm}}
\toprule
\rowcolor{tableheader}
\textbf{Task (Num.)} & \textbf{Subtask (Num.)} & \textbf{Sensor} & \textbf{Method} & \textbf{Source Datasets} \\
\midrule
\multirow[t]{6}{*}{\makecell[l]{Captioning\\(7.5M)}}
& \makecell[l]{Image Captioning\\(5.2M)} & O, S, M, T, C & F, B & NWPU-Captions\cite{cheng2022nwpu}, RSICD\cite{lu2017exploring}, UCM-Captions\cite{lu2017exploring}, Sydney-Captions\cite{lu2017exploring}, RemoteCLIP\cite{liu2024remoteclip}, HQRS-IT\cite{he2025enhancing}, GeoText\cite{chu2024towards}, RSCap\cite{hu2023rsgpt}, RSD46-WHU\cite{long2017accurate}, DIOR\cite{li2020object}, WHU-RS19\cite{balestra2025whu}, XLRS-Bench\cite{wang2025xlrs}, IRGPT\cite{cao2025irgpt}, SAR-Text\cite{he2025sar}, SARCap\cite{jiang2026sarclip}, EarthDial-Instruct\cite{soni2025earthdial}, SARLANG-1M\cite{wei2026sarlang}, QXS-SAROPT\cite{huang2021qxs}, MultiResSAR\cite{zhang2025multi}, WHU-OPT-SAR\cite{wei2024mgfnet}, ChatEarthNet\cite{yuan2024chatearthnet}, EarthMind-Instruct\cite{earthmind2025}, FIT-RS\cite{luo2024skysensegpt}, LEVIR-MCI\cite{liu2024change}, WHU-CDC\cite{shi2024whucdc}, SECOND\cite{yang2020semantic}, S2Looking\cite{shen2021s2looking}, SYSU-CD\cite{shi2021deeply}, HIUCD\cite{tian2020hi}, OpenSuWu\cite{shi2023openWUSU}, TAMMS\cite{guo2025tamms} \\
\rowcolor{gray!6}
& \makecell[l]{Region Captioning\\(1.2M)} & O, S, I, T & F, B & DIOR-RSVG\cite{zhan2023rsvg}, OPT-RSVG\cite{li2024language}, RSVG\cite{zhan2023rsvg}, RSVG-HR\cite{lan2024language}, RSSVG\cite{chen2025vgrss}, SARVG\cite{chen2025vgrss}, VRSBench\cite{li2024vrsbench}, EarthDial-Instruct\cite{soni2025earthdial}, TEOChatlas\cite{irvin2024teochat}, GeoChat-Instruct\cite{kuckreja2023geochat}, XLRS-Bench\cite{wang2025xlrs}, FIT-RS\cite{luo2024skysensegpt}, GeoText\cite{chu2024towards}, DE-Dataset\cite{li2025describeearth} \\
& \makecell[l]{Change Captioning\\(0.7M)} & O, S, C, T & F, G & ChangeChat\cite{deng2025changechat}, SYSU-CD\cite{shi2021deeply}, EarthDial-Instruct\cite{soni2025earthdial}, SECOND\cite{yang2020semantic}, SECOND-CC\cite{karaca2025robust}, EBD\cite{chen2025rscc}, GeoLLaVA\cite{elgendy2024geollava}, DisasterM3\cite{wang2025disasterm3}, TEOChatlas\cite{irvin2024teochat}, LEVIR-MCI\cite{liu2024change}, WHU-CDC\cite{shi2024whucdc}, DVL-Instruct\cite{xuan2025dynamicvl}, Landsat30AU\cite{ma2026landsat30} \\
\rowcolor{gray!6}
& \makecell[l]{Grounded Description\\(0.3M)} & O, S, I, T & F, B & FIT-RS\cite{luo2024skysensegpt}, GeoChat-Instruct\cite{kuckreja2023geochat}, HIT-UAV\cite{suo2023hit}, TEOChatlas\cite{irvin2024teochat} \\
& \makecell[l]{Video Captioning\\(37K)} & V & F, G & CapERA\cite{bashmal2023capera}, Sat-MTB-MLSC\cite{guo2024satmtbmlsc}, Sat-SOT\cite{zhao2022satsot}, VISO\cite{zhao2022satsot} \\
\rowcolor{gray!6}
& \makecell[l]{Temporal Captioning\\(0.8K)} & T & F & DVL-Instruct\cite{xuan2025dynamicvl} \\
\midrule
\multirow[t]{3}{*}{\makecell[l]{VQA\\(9.9M)}}
& \makecell[l]{Image VQA\\(8.4M)} & O, S, I, M, C & F, B & RSVQA-HR\cite{lobry2020rsvqa}, RSVQA-LR\cite{lobry2020rsvqa}, EarthVQA\cite{wang2024earthvqa}, FloodNet\cite{rahnemoonfar2021floodnet}, RescueNet\cite{rahnemoonfar2023rescuenet}, OSVQA\cite{zhao2025text}, RSVLM\cite{zi2025rsvlm}, GeoLLaVA-8K\cite{wanggeollava}, RS-GPT4V\cite{xu2024rs}, GeoChat-Instruct\cite{kuckreja2023geochat}, FIT-RS\cite{luo2024skysensegpt}, VHM-RS\cite{pang2025vhm}, LRS-GRO\cite{liu2025zoomearth}, FarmSeg\cite{tao2025large}, MME-RealWorld\cite{zhangmme}, XLRS-Bench\cite{wang2025xlrs}, SARLANG-1M\cite{wei2026sarlang}, GeoLlama\cite{li2024geo}, EarthDial-Instruct\cite{soni2025earthdial}, EarthMind-Instruct\cite{earthmind2025} \\
\rowcolor{gray!6}
& \makecell[l]{Temporal VQA\\(1.4M)} & O, S, C, T & F, G & DVL-Instruct\cite{xuan2025dynamicvl}, DisasterM3\cite{wang2025disasterm3}, LEVIR-MCI\cite{liu2024change}, SECOND\cite{yang2020semantic}, TEOChatlas\cite{irvin2024teochat}, WHU-CDC\cite{shi2024whucdc}, EarthDial-Instruct\cite{soni2025earthdial}, Landsat30AU\cite{ma2026landsat30} \\
& \makecell[l]{Video VQA\\(0.1M)} & V & F, G & ERA\cite{mou2020era}, Sat-MTB-MLSC\cite{guo2024satmtbmlsc}, Sat-SOT\cite{zhao2022satsot}, VISO\cite{zhao2022satsot} \\
\midrule
\multirow[t]{4}{*}{\makecell[l]{Classification\\(2.7M)}}
& \makecell[l]{Image Classification\\(2.4M)} & O, S, M, C & F, B & NWPU-RESISC45\cite{cheng2017remote}, EuroSAT\cite{helber2018introducing}, PatternNet\cite{zhou2018patternnet}, FMoW\cite{christie2018functional}, RSD46-WHU\cite{long2017accurate}, VHM-RS\cite{pang2025vhm}, WHU-RS19\cite{balestra2025whu}, FIT-RS\cite{luo2024skysensegpt}, GeoChat-Instruct\cite{kuckreja2023geochat}, SARLANG-1M\cite{wei2026sarlang}, EarthDial-Instruct\cite{soni2025earthdial}, EarthMind-Instruct\cite{earthmind2025} \\
\rowcolor{gray!6}
& \makecell[l]{Region Classification\\(0.2M)} & O, S & F & DIOR-RSVG\cite{zhan2023rsvg}, OPT-RSVG\cite{li2024language}, RSSVG\cite{chen2025vgrss}, RSVG\cite{zhan2023rsvg}, RSVG-HR\cite{lan2024language}, SARVG\cite{chen2025vgrss} \\
& \makecell[l]{Temporal Classification\\(0.1M)} & T & F & LEVIR-MCI\cite{liu2024change}, TEOChatlas\cite{irvin2024teochat}, WHU-CDC\cite{shi2024whucdc}, EarthDial-Instruct\cite{soni2025earthdial} \\
\rowcolor{gray!6}
& \makecell[l]{Video Classification\\(1.5K)} & V & F & ERA\cite{mou2020era}, Sat-MTB-MLSC\cite{guo2024satmtbmlsc} \\
\midrule
\multirow[t]{1}{*}{\makecell[l]{Detection\\(6.5M)}}
& \makecell[l]{Detection\\(6.5M)} & O, S, I, T & F & DIOR\cite{li2020object}, DOTA\cite{xia2018dota}, xView\cite{lam2018xview}, VisDrone\cite{cao2021visdrone}, FAIR1M\cite{sun2022fair1m}, HRRSD\cite{zhang2019hrrsd}, NWPU-RESISC45\cite{cheng2017remote}, RSOD\cite{long2017accurate}, UCAS-AOD\cite{zhu2015orientation}, ShipDataset\cite{sdu2020oceanicship}, WHU-Building\cite{luo2023diverse}, FIT-RS\cite{luo2024skysensegpt}, SARDet\cite{li2024sardet}, HRSID\cite{wei2020hrsid}, SSDD\cite{zhang2021sar}, SAR-AIRcraft\cite{zhirui2023sar}, SRSDD\cite{lei2021srsdd}, HIT-UAV\cite{suo2023hit}, Sea-shipping\cite{infrared2021seashipping}, Infrared-security\cite{infrared2021security}, EarthDial-Instruct\cite{soni2025earthdial}, Double-light-vehicle\cite{infrared2021doublevehicle}, Aerial-mancar\cite{infrared2021aerialmancar}, TEOChatlas\cite{irvin2024teochat} \\
\midrule
\multirow[t]{3}{*}{\makecell[l]{Visual Grounding\\(0.7M)}}
& \makecell[l]{Image Grounding\\(0.6M)} & O, S, I & F, B & DIOR-RSVG\cite{zhan2023rsvg}, OPT-RSVG\cite{li2024language}, RSVG\cite{zhan2023rsvg}, RSVG-HR\cite{lan2024language}, RSSVG\cite{chen2025vgrss}, VRSBench\cite{li2024vrsbench}, DVGBench\cite{zhou2026dvgbench}, UrBench\cite{zhou2025urbench}, XLRS-Bench\cite{wang2025xlrs}, FIT-RS\cite{luo2024skysensegpt}, GeoChat-Instruct\cite{kuckreja2023geochat}, SARVG\cite{chen2025vgrss}, EarthDial-Instruct\cite{soni2025earthdial} \\
\rowcolor{gray!6}
& \makecell[l]{Temporal Grounding\\(7.4K)} & O, T & F & TEOChatlas\cite{irvin2024teochat}, EarthDial-Instruct\cite{soni2025earthdial} \\
& \makecell[l]{Video Grounding\\(0.07M)} & V & F & UAVSVG\cite{zhan2025does} \\
\midrule
\multirow[t]{2}{*}{\makecell[l]{Segmentation\\(3.6M)}}
& \makecell[l]{Semantic Segmentation\\(2.0M)} & O & F & CrowdAI\cite{adimoolam2023efficient}, LoveDA\cite{wang2021loveda}, OpenEarthMap\cite{xia2023openearthmap}, Potsdam\cite{isprs2021potsdam}, WHU-Mixer\cite{luo2023diverse}, WHU-Building\cite{luo2023diverse}, LEVIR-MCI\cite{liu2024change}, WHU-CDC\cite{shi2024whucdc} \\
\rowcolor{gray!6}
& \makecell[l]{Referring/Reasoning\\Segmentation\\(1.6M)} & O, T & F & GeoPixInstruct\cite{ou2025geopix}, RefGeo\cite{zhou2024geoground}, EarthReason\cite{li2025segearth}, RRSIS-D\cite{liu2024rotated}, RISBench\cite{dong2024cross}, RemoteSAM\cite{yao2025remotesam}, NWPU-Ref\cite{yang2025large}, RefSegRS\cite{yuan2023rrsis} \\
\midrule
\multirow[t]{1}{*}{\makecell[l]{Change Detection\\(0.2M)}}
& \makecell[l]{Change Detection\\(0.2M)} & O, T & F & LEVIR-MCI\cite{liu2024change}, WHU-CDC\cite{shi2024whucdc}, S2Looking\cite{shen2021s2looking}, UAV-BCD\cite{ying2023uav}, DVL-Instruct\cite{xuan2025dynamicvl}, TEOChatlas\cite{irvin2024teochat} \\
\midrule
\multirow[t]{4}{*}{\makecell[l]{Visual Prompting\\(3.1M)}}
& \makecell[l]{VP-Caption\\(0.7M)} & O, S, I, C, T & F, G & DIOR-RSVG\cite{zhan2023rsvg}, OPT-RSVG\cite{li2024language}, RSVG\cite{zhan2023rsvg}, RSVG-HR\cite{lan2024language}, RSSVG\cite{chen2025vgrss}, SARVG\cite{chen2025vgrss}, FIT-RS\cite{luo2024skysensegpt}, GeoChat-Instruct\cite{kuckreja2023geochat}, DIOR\cite{li2020object}, DOTA\cite{xia2018dota}, HIT-UAV\cite{suo2023hit}, Infrared-security\cite{infrared2021security}, SARDet\cite{li2024sardet}, LEVIR-MCI\cite{liu2024change}, WHU-CDC\cite{shi2024whucdc} \\
\rowcolor{gray!6}
& \makecell[l]{VP-Classification\\(0.7M)} & O, S, I, C, T & F & DIOR\cite{li2020object}, DOTA\cite{xia2018dota}, FAIR1M\cite{sun2022fair1m}, HRRSD\cite{zhang2019hrrsd}, HRSC2016\cite{chen2022mssdet}, RSD-GOD\cite{zhuang2019single}, RSOD\cite{long2017accurate}, UCAS-AOD\cite{zhu2015orientation}, SARDet\cite{li2024sardet}, MSAR\cite{chen2022msar}, SAR-AIRcraft\cite{zhirui2023sar}, SARShip\cite{zhang2021sar}, SSDD\cite{zhang2021sar}, HIT-UAV\cite{suo2023hit}, Sea-shipping\cite{infrared2021seashipping}, Infrared-security\cite{infrared2021security}, Double-light-vehicle\cite{infrared2021doublevehicle}, LEVIR-MCI\cite{liu2024change}, WHU-CDC\cite{shi2024whucdc} \\
& \makecell[l]{VP-VQA\\(1.5M)} & O, S, I, C, T & F, G & DIOR\cite{li2020object}, Infrared-security\cite{infrared2021security}, SARDet\cite{li2024sardet}, LEVIR-MCI\cite{liu2024change}, WHU-CDC\cite{shi2024whucdc} \\
\rowcolor{gray!6}
& \makecell[l]{VP-Relationship\\(0.1M)} & O, S, I, C, T & F, G & DIOR\cite{li2020object}, DOTA\cite{xia2018dota}, HIT-UAV\cite{suo2023hit}, Infrared-security\cite{infrared2021security}, SARDet\cite{li2024sardet}, LEVIR-MCI\cite{liu2024change}, WHU-CDC\cite{shi2024whucdc} \\
\midrule
\multirow[t]{1}{*}{\makecell[l]{Relation\\(0.5M)}}
& \makecell[l]{Image/Region\\Relation\\(0.5M)} & O & B & FIT-RS\cite{luo2024skysensegpt} \\
\midrule
\rowcolor{oursrow}
\best{9 tasks (34M+)} & \best{25 subtasks (34M+)} & \best{7 types} & --- & --- \\
\bottomrule
\end{tabular}
}
\end{table*}

\textbf{Captioning} (7.5M).
Captioning builds multi-granularity language descriptions across six subtasks: image captioning (e.g., RSICD, SARLANG-1M, etc., 5.2M), region captioning (e.g., DIOR-RSVG, VRSBench, etc.), grounded description (e.g., FIT-RS, GeoChat-Instruct), change captioning, and video and temporal captioning. GPT-4o expands brief annotations from LEVIR-MCI, WHU-CDC, SECOND into detailed change narratives and synthesizes video descriptions from ERA, Sat-MTB key frames.

\textbf{Visual Question Answering} (9.9M).
VQA comprises image VQA across five modalities (e.g., RSVQA-HR, SARLANG-1M, etc., 8.4M), temporal VQA (e.g., TEOChatlas, DVL-Instruct, etc., 1.4M), and video VQA from frame-sampled RS videos (e.g., ERA, VISO, etc.). GPT-4o synthesizes diverse question types for temporal and video subsets.

\textbf{Classification} (2.7M).
Classification spans image, region, temporal, and video levels. Multispectral classification (e.g., BigEarthNet, EuroSAT-MS, etc.) fills a gap absent from prior RS instruction datasets. Region classification reuses grounding annotations for dual localization-classification utility. Temporal classification converts change detection annotations (e.g., LEVIR-MCI, WHU-CDC) into scene-state samples.

\textbf{Detection} (6.5M).
Detection integrates 24 datasets across optical (e.g., DIOR, DOTA, etc.), SAR (e.g., SARDet, SSDD, etc.), infrared (HIT-UAV), and temporal modalities. Horizontal boxes, oriented boxes, and center points are unified through the same coordinate token encoding.

\textbf{Visual Grounding} (0.7M).
Visual grounding maps free-form referring expressions to spatial coordinates across image (e.g., DIOR-RSVG, VRSBench, etc.), temporal (e.g., TEOChatlas, xBD), and video (e.g., UAVSVG) subtasks. Each annotation is reused across tasks to produce region captioning and region classification samples.

\textbf{Segmentation} (3.6M).
Segmentation covers semantic segmentation (e.g., LoveDA, OpenEarthMap, etc.) and referring segmentation (e.g., GeoPixInstruct, RRSIS-D, etc.). R-RLE encoding (Section~\ref{sec:spatial_format}) compresses arbitrary-shape masks into compact token sequences without additional output heads.

\textbf{Change Detection} (0.2M).
Change detection produces spatial change masks that answer ``where did change occur?'', with R-RLE masks for pixel-level localization and bounding boxes for object-level perception, covering building changes (e.g., LEVIR-MCI, WHU-CDC), post-disaster damage (e.g., S2Looking, TEOChatlas), and land-use transitions (e.g., UAV-BCD, DVL-Instruct).

\textbf{Visual Prompting} (3.1M).
Visual prompting allows users to mark regions via points, boxes, or ellipses on a separate blank image alongside the original as dual-image input, supporting VP-caption, VP-classification, VP-VQA, and VP-relationship across optical, SAR, infrared, and temporal modalities.

\textbf{Relation Reasoning} (0.5M).
Relation reasoning builds structured understanding of inter-object spatial and functional associations, sourced from the complex understanding subset of FIT-RS~\cite{luo2024skysensegpt}.

\subsection{Data Integrity}
\label{sec:decontamination}

Test splits of downstream evaluation benchmarks are excluded from all stages of \datasetname{} data collection, conversion, and synthesis.

\begin{table*}[!t]
\centering
\caption{\small Classification accuracy (\%) on scene classification, xBD disaster assessment, and FMoW temporal classification benchmarks. \best{Bold red} denotes the best result and \second{underlined blue} the second best.}
\label{tab:classification}
\setlength{\tabcolsep}{5pt}
\renewcommand{\arraystretch}{0.75}
\resizebox{\textwidth}{!}{%
\scriptsize
\begin{tabular}{l|ccc|cccccc|c}
\toprule
\rowcolor{tableheader}
& \multicolumn{3}{c|}{\textbf{Scene Classification} \small{(1-image)}}
& \multicolumn{6}{c|}{\makebox[0pt][c]{\textbf{Temporal Classification} \small{(multi-images)}}} & \\
\cmidrule(lr){2-4} \cmidrule(lr){5-11}
\rowcolor{tableheader}
& \textbf{UCMerced} & \textbf{WHU-RS19} & \textbf{BEN-RGB}
& \multicolumn{6}{c|}{\textbf{xBD} \small{(2-images)}}
& \textbf{FMoW} \small{(4-images)} \\
\cmidrule(lr){2-4} \cmidrule(lr){5-9} \cmidrule(lr){10-10} \cmidrule(lr){11-11}
\rowcolor{tableheader}
& -- & -- & --
& \textbf{Region Cls.-1} & \textbf{Region Cls.-2} & \textbf{DisasterType} & \textbf{BuildingDmg} & \textbf{DmgCount} & \cellcolor{avgcol}\textbf{Avg}
& -- \\
\cmidrule(lr){2-4} \cmidrule(lr){5-10} \cmidrule(lr){11-11}
\rowcolor{tableheader}
\textbf{Model}
& \textbf{Acc} & \textbf{Acc} & \textbf{Recall}
& \textbf{Acc} & \textbf{Acc} & \textbf{Acc} & \textbf{Acc} & \textbf{Acc} & \cellcolor{avgcol}\textbf{Acc}
& \textbf{Acc} \\
\midrule
GPT-4o        & 88.76 & 91.14 & 49.00 & 51.68 & 71.62 & 67.95 & 75.45 & \best{70.41} & \cellcolor{avgcol}67.42 & 21.43 \\
\rowcolor{gray!6}
InternVL2 (8B)  & 58.23 & 79.30 & 19.73 & 14.39 & 58.33 & 51.44 & 61.52 & 51.12 & \cellcolor{avgcol}47.36 & 21.04 \\
Qwen2.5-VL (3B) & 60.86 & 78.21 & 24.75 & --    & --    & --    & --    & --    & \cellcolor{avgcol}-- & 34.36 \\
\rowcolor{gray!6}
GeoChat (7B)       & 84.43 & 80.09 & 20.35 & 25.30 & 57.65 & 53.32 & 52.19 & 49.51 & \cellcolor{avgcol}47.59 & 59.20 \\
SkySenseGPT (7B)   & 85.00 & 93.16 & --    & --    & --    & --    & --    & --    & \cellcolor{avgcol}-- & --    \\
\rowcolor{gray!6}
VHM (7B)           & 89.29 & 91.84 & --    & --    & --    & --    & --    & --    & \cellcolor{avgcol}-- & --    \\
EarthDial (4B)     & \best{92.42} & \second{96.21} & \second{73.03} & \second{53.70} & \second{83.09} & \second{96.37} & \second{82.85} & 54.01 & \cellcolor{avgcol}\second{74.00} & \second{70.03} \\
\midrule
\rowcolor{oursrow}
\textbf{\modelname{} (2B)} & \second{91.83} & \best{97.31} & \best{78.03} & \best{76.33} & \best{84.62} & \best{98.07} & \best{85.64} & \second{59.81} & \cellcolor{avgcol}\best{80.89} & \best{73.63} \\
\bottomrule
\multicolumn{11}{l}{\scriptsize BEN-RGB = BigEarthNet RGB (multi-label), Region Cls. = Region Classification, DmgCount = Damage Count.}
\end{tabular}}
\end{table*}

\begin{table*}[!t]
\centering
\setlength{\tabcolsep}{5pt}
\renewcommand{\arraystretch}{0.75}
\caption{\small Captioning comparison on change, image, temporal, and region captioning testsets. Following EarthDial, R-1, R-L, and METEOR computed using NLTK\@. \best{Bold red} denotes the best result and \second{underlined blue} the second best.}
\label{tab:captioning_comprehensive}
\resizebox{\textwidth}{!}{%
\scriptsize
\begin{tabular}{l|ccc|ccc||ccc|ccc||ccc||ccc}
\toprule
\rowcolor{tableheader}
& \multicolumn{6}{c||}{\textbf{Change Captioning}} & \multicolumn{6}{c||}{\textbf{Image Captioning}} & \multicolumn{3}{c||}{\textbf{Temporal Captioning}} & \multicolumn{3}{c}{\textbf{Region Captioning}} \\
\cmidrule(lr){2-7} \cmidrule(lr){8-13} \cmidrule(lr){14-16} \cmidrule(lr){17-19}
\rowcolor{tableheader}
\textbf{Model} & \multicolumn{3}{c|}{\textbf{LEVIR-MCI}} & \multicolumn{3}{c||}{\textbf{MUDS}} & \multicolumn{3}{c|}{\textbf{RSICD}} & \multicolumn{3}{c||}{\textbf{RSITMD(ZS)}} & \multicolumn{3}{c||}{\textbf{xBD}} & \multicolumn{3}{c}{\textbf{SAR-Ship}} \\
\cmidrule(lr){2-4} \cmidrule(lr){5-7} \cmidrule(lr){8-10} \cmidrule(lr){11-13} \cmidrule(lr){14-16} \cmidrule(lr){17-19}
\rowcolor{tableheader}
& \textbf{R-1} & \textbf{R-L} & \textbf{MT} & \textbf{R-1} & \textbf{R-L} & \textbf{MT} & \textbf{R-1} & \textbf{R-L} & \textbf{MT} & \textbf{R-1} & \textbf{R-L} & \textbf{MT} & \textbf{R-1} & \textbf{R-L} & \textbf{MT} & \textbf{R-1} & \textbf{R-L} & \textbf{MT} \\
\midrule
GPT-4o         & 10.33 & 8.40  & 22.05 & 14.18 & 11.02 & 20.92 & 20.53 & 15.59 & 26.03 & 18.31 & 14.22 & 24.83 & 14.21 & 10.35 & 19.52 & 7.49  & 7.24  & 7.07  \\
\rowcolor{gray!6}
InternVL2 (4B)   & 8.88  & 7.43  & 22.14 & 10.25 & 7.90  & 17.73 & 0.00  & 0.00  & 0.00  & 0.00  & 0.00  & 0.00  & --    & --    & --    & --    & --    & --    \\
InternVL2 (8B)   & --    & --    & --    & --    & --    & --    & 21.59 & 16.13 & 28.17 & 18.91 & 14.65 & 26.02 & 13.89 & 10.37 & 14.92 & 9.67  & 8.67  & 8.19  \\
\rowcolor{gray!6}
Qwen2.5-VL (3B)  & 12.27 & 10.11 & 26.11 & 12.13 & 9.30  & 18.22 & 21.37 & 16.42 & 26.53 & 18.79 & 15.02 & 25.05 & --    & --    & --    & --    & --    & --    \\
GeoChat (7B)        & 17.15 & \best{35.42} & 12.35 & 12.28 & 12.23 & 15.98 & 13.48 & 11.59 & 12.39 & 13.41 & 11.50 & 12.33 & 14.18 & 10.67 & 12.20 & 57.15 & 57.15 & 52.20 \\
\rowcolor{gray!6}
EarthDial (4B)      & \second{33.78} & 30.47 & \second{74.80} & \second{28.16} & \second{24.03} & \second{33.56} & \best{33.77} & \best{27.61} & \best{56.18} & \second{26.74} & \second{21.72} & \best{34.06} & \second{87.26} & \second{87.26} & \second{88.53} & \second{63.10} & \second{63.10} & \second{54.83} \\
\midrule
\rowcolor{oursrow}
\textbf{\modelname{} (2B)} & \best{33.83} & \second{30.73} & \best{76.45} & \best{33.98} & \best{26.93} & \best{37.40} & \second{30.38} & \second{24.13} & \second{32.39} & \best{27.57} & \best{22.32} & \second{27.55} & \best{92.33} & \best{92.31} & \best{93.63} & \best{78.63} & \best{78.63} & \best{60.18} \\
\bottomrule
\end{tabular}}
\end{table*}

\begin{table*}[!t]
\centering
\caption{\small Visual grounding comparison on five testsets. P@0.5 denotes precision at IoU threshold 0.5. \best{Bold red} denotes the best result and \second{underlined blue} the second best.}
\label{tab:visual_grounding}
\setlength{\tabcolsep}{11pt}
\renewcommand{\arraystretch}{0.65}
\resizebox{\textwidth}{!}{
\scriptsize
\begin{tabular}{l|cc|cc|cc|cc|cc}
\toprule
\rowcolor{tableheader}
& \multicolumn{2}{c|}{\textbf{DIOR-RSVG}} & \multicolumn{2}{c|}{\textbf{OPT-RSVG}} & \multicolumn{2}{c|}{\textbf{VRSBench-VG}} & \multicolumn{2}{c|}{\textbf{RSVG}} & \multicolumn{2}{c}{\textbf{RSVG-HR}} \\
\cmidrule(lr){2-3} \cmidrule(lr){4-5} \cmidrule(lr){6-7} \cmidrule(lr){8-9} \cmidrule(lr){10-11}
\rowcolor{tableheader}
\textbf{Model} & \textbf{P@0.5} & \textbf{mIoU} & \textbf{P@0.5} & \textbf{mIoU} & \textbf{P@0.5} & \textbf{mIoU} & \textbf{P@0.5} & \textbf{mIoU} & \textbf{P@0.5} & \textbf{mIoU} \\
\midrule
\rowcolor{gray!20}
\textcolor{gray}{SOTA Specialist} & \textcolor{gray}{84.35} & \textcolor{gray}{75.10} & \textcolor{gray}{\second{86.51}} & \textcolor{gray}{\best{76.47}} & \textcolor{gray}{82.98} & \textcolor{gray}{70.28} & \textcolor{gray}{\best{84.48}} & \textcolor{gray}{\best{74.62}} & \textcolor{gray}{\best{87.40}} & \textcolor{gray}{\best{75.84}} \\
\midrule
Qwen2.5-VL (3B) & 36.3 & 34.34 & 24.4 & 25.50 & 45.2 & 42.45 & 1.0 & 7.24 & 29.4 & 32.1 \\
\rowcolor{gray!6}
GLM-4.1V-Thinking (9B) & 59.6 & 57.41 & 58.7 & 54.13 & 63.8 & 60.69 & 43.0 & 42.27 & 30.7 & 29.8 \\
SkySenseGPT (7B) & 60.8 & 53.18 & - & - & 63.5 & 54.60 & 39.5 & 38.54 & - & - \\
EarthDial (4B) & 46.1 & 39.46 & 53.6 & 46.49 & 14.4 & 13.00 & 42.0 & 38.49 & - & - \\
\rowcolor{gray!6}
GeoViS (3B) & 79.8 & 72.60 & 70.3 & 61.50 & 68.5 & 59.20 & - & - & 51.5 & 46.5 \\
RSThinker (9B) & \second{93.1} & \best{89.02} & 63.8 & 64.31 & \second{90.4} & \best{80.79} & 64.0 & \second{59.74} & - & - \\
\midrule
\rowcolor{oursrow}
\textbf{\modelname{} (2B)} & \best{94.41} & \second{87.96} & \best{87.52} & \second{75.38} & \best{90.77} & \second{77.04} & \second{65.06} & 52.95 & \second{82.36} & \second{66.86} \\
\bottomrule
\end{tabular}
}
\end{table*}

\begin{table*}[t]
\centering
\caption{\small Comparison on VRSBench-VQA, RSVQA-HR, and RSVQA-LR testsets. VRSBench-VQA evaluation uses GPT-4o-mini semantic matching (official protocol). Avg denotes the mean accuracy across question categories. RSVQA-HR and RSVQA-LR metrics are accuracy (\%). \best{Bold red} denotes the best result and \second{underlined blue} the second best.}
\label{tab:rsvqa}
\setlength{\tabcolsep}{6pt}
\renewcommand{\arraystretch}{0.68}
\resizebox{\linewidth}{!}{
\scriptsize
\begin{tabular}{l|cccccccc|ccc|cccc}
\toprule
\rowcolor{tableheader}
& \multicolumn{8}{c|}{\textbf{VRSBench-VQA}} & \multicolumn{3}{c|}{\textbf{RSVQA-HR}} & \multicolumn{4}{c}{\textbf{RSVQA-LR}} \\
\cmidrule(lr){2-9} \cmidrule(lr){10-12} \cmidrule(lr){13-16}
\rowcolor{tableheader}
\textbf{Model} & \textbf{Cat.} & \textbf{Exist.} & \textbf{Pos.} & \textbf{Quant.} & \textbf{Scene} & \textbf{Color} & \textbf{Image} & \cellcolor{avgcol}\textbf{Avg} & \textbf{Pres.} & \textbf{Comp.} & \cellcolor{avgcol}\textbf{Avg} & \textbf{Pres.} & \textbf{Comp.} & \textbf{R/U} & \cellcolor{avgcol}\textbf{Avg} \\
\midrule
Gemini-2.0-Flash  & 44.03 & 86.11 & 43.97 & 46.00 & 60.56 & 56.96 & \best{95.83} & \cellcolor{avgcol}61.92 & 56.94 & 42.96 & \cellcolor{avgcol}49.95 & -- & -- & -- & \cellcolor{avgcol}-- \\
\rowcolor{gray!6}
MiniGPT-v2 (7B)        & 25.37 & 56.25 & 20.69 & 44.00 & 45.07 & 36.71 & 33.33 & \cellcolor{avgcol}37.35 & 48.95 & 52.95 & \cellcolor{avgcol}50.95 & 55.16 & 55.22 & 39.00 & \cellcolor{avgcol}49.79 \\
\rowcolor{gray!6}
Kimi-VL-Thinking (16B)  & 47.01 & 87.50 & 46.55 & \best{74.67} & 71.83 & \second{65.82} & 90.23 & \cellcolor{avgcol}69.09 & 63.94 & 77.91 & \cellcolor{avgcol}70.93 & -- & -- & -- & \cellcolor{avgcol}-- \\
SkySenseGPT (7B)       & 57.46 & 84.03 & 44.83 & 38.00 & 53.52 & 16.46 & 45.83 & \cellcolor{avgcol}48.59 & 47.95 & 78.93 & \cellcolor{avgcol}63.44 & -- & -- & -- & \cellcolor{avgcol}-- \\
\rowcolor{gray!6}
SkyEyeGPT (7B)         & -- & -- & -- & -- & -- & -- & -- & \cellcolor{avgcol}-- & \second{83.50} & 80.28 & \cellcolor{avgcol}\second{81.89} & 88.93 & 88.63 & 75.00 & \cellcolor{avgcol}84.19 \\
TEOChat (7B)           & -- & -- & -- & -- & -- & -- & -- & \cellcolor{avgcol}-- & 67.50 & 81.10 & \cellcolor{avgcol}74.30 & 91.70 & \second{92.70} & \second{94.00} & \cellcolor{avgcol}92.80 \\
\rowcolor{gray!6}
EarthDial (4B)         & 51.49 & 47.22 & 36.21 & 41.33 & 36.62 & 11.39 & 50.00 & \cellcolor{avgcol}39.18 & 58.89 & \second{83.11} & \cellcolor{avgcol}71.0 & \best{92.58} & \best{92.75} & \second{94.00} & \cellcolor{avgcol}\best{93.11} \\
RSThinker (9B)         & \second{82.84} & \best{92.36} & \second{68.97} & 56.67 & \second{73.24} & 64.33 & \second{92.87} & \cellcolor{avgcol}\second{75.90} & 66.95 & 78.98 & \cellcolor{avgcol}72.97 & -- & -- & -- & \cellcolor{avgcol}-- \\
\midrule
\rowcolor{oursrow}
\textbf{\modelname{} (2B)} & \best{88.70} & \second{87.96} & \best{73.46} & \second{67.48} & \best{78.73} & \best{73.52} & 92.38 & \cellcolor{avgcol}\best{80.32} & \best{84.70} & \best{88.02} & \cellcolor{avgcol}\best{86.36} & \second{91.83} & 92.39 & \best{94.50} & \cellcolor{avgcol}\second{92.91} \\
\bottomrule
\end{tabular}}
\end{table*}

\begin{table*}[!t]
\centering
\caption{\small Referred object detection comparison on four referred object detection testsets. Detection metrics are Precision@IoU$\geq$0.5 (\%).}
\label{tab:earthdial_comprehensive}
\footnotesize
\setlength{\tabcolsep}{2.2pt}
\renewcommand{\arraystretch}{0.72}
\resizebox{\textwidth}{!}{%
\begin{tabular}{l|cccccc|cccccc|cccccc|cccccc}
\toprule
\rowcolor{tableheader}
& \multicolumn{6}{c|}{\textbf{GeoChat-Instruct}} & \multicolumn{6}{c|}{\textbf{NWPU-VHR-10}} & \multicolumn{6}{c|}{\textbf{SwimmingPool (ZS)}} & \multicolumn{6}{c}{\textbf{UrbanTreeCrown (ZS)}} \\
\cmidrule(lr){2-7} \cmidrule(lr){8-13} \cmidrule(lr){14-19} \cmidrule(lr){20-25}
\rowcolor{tableheader}
\textbf{Model} & \textbf{Sm} & \textbf{Md} & \textbf{Lg} & \textbf{Sg} & \textbf{Mt} & \cellcolor{avgcol}\textbf{Avg} & \textbf{Sm} & \textbf{Md} & \textbf{Lg} & \textbf{Sg} & \textbf{Mt} & \cellcolor{avgcol}\textbf{Avg} & \textbf{Sm} & \textbf{Md} & \textbf{Lg} & \textbf{Sg} & \textbf{Mt} & \cellcolor{avgcol}\textbf{Avg} & \textbf{Sm} & \textbf{Md} & \textbf{Lg} & \textbf{Sg} & \textbf{Mt} & \cellcolor{avgcol}\textbf{Avg} \\
\midrule
InternVL2 (4B) & 6.30 & 24.37 & 37.38 & 24.96 & 11.72 & \cellcolor{avgcol}20.95 & 7.10 & 12.68 & \second{25.48} & 22.96 & 8.10 & \cellcolor{avgcol}15.26 & 0.60 & 6.60 & 8.90 & 4.50 & 0.87 & \cellcolor{avgcol}4.29 & -- & 3.17 & 13.41 & 5.90 & 3.10 & \cellcolor{avgcol}-- \\
\rowcolor{gray!6}
InternVL2 (8B) & 7.20 & 23.76 & 31.99 & 25.77 & 9.30 & \cellcolor{avgcol}19.60 & 4.26 & 11.85 & 20.72 & 21.66 & 5.86 & \cellcolor{avgcol}12.87 & 0.30 & 4.70 & 18.27 & 7.60 & 0.51 & \cellcolor{avgcol}6.28 & 0.60 & 3.99 & 17.10 & 7.90 & 3.94 & \cellcolor{avgcol}6.71 \\
GeoChat (7B) & 2.90 & 13.60 & 21.70 & 16.00 & 4.30 & \cellcolor{avgcol}11.70 & 2.50 & 3.20 & 14.70 & 13.23 & 1.90 & \cellcolor{avgcol}7.11 & -- & 3.10 & 7.30 & 1.20 & 0.60 & \cellcolor{avgcol}-- & -- & 1.80 & 8.90 & 2.90 & 3.10 & \cellcolor{avgcol}-- \\
\rowcolor{gray!6}
EarthDial (4B) & \second{11.43} & \second{31.76} & \second{39.07} & \second{34.29} & \second{13.41} & \cellcolor{avgcol}\second{25.99} & \second{11.66} & \second{14.21} & 23.12 & \second{25.37} & \second{8.90} & \cellcolor{avgcol}\second{16.65} & \second{1.04} & \second{7.40} & \second{24.90} & \second{8.40} & \second{1.04} & \cellcolor{avgcol}\second{8.56} & \second{1.10} & \second{7.01} & \second{25.67} & \second{11.13} & \second{6.70} & \cellcolor{avgcol}\second{10.32} \\
\midrule
\rowcolor{oursrow}
\textbf{\modelname{} (2B)} & \best{28.48} & \best{47.87} & \best{49.38} & \best{52.03} & \best{25.37} & \cellcolor{avgcol}\best{40.63} & \best{50.63} & \best{59.41} & \best{67.02} & \best{62.19} & \best{56.89} & \cellcolor{avgcol}\best{59.23} & \best{5.54} & \best{22.93} & \best{43.12} & \best{24.14} & \best{4.35} & \cellcolor{avgcol}\best{20.02} & \best{4.62} & \best{16.70} & \best{39.42} & \best{24.08} & \best{13.83} & \cellcolor{avgcol}\best{19.73} \\
\bottomrule
\multicolumn{25}{l}{\scriptsize Sm=Small, Md=Medium, Lg=Large, Sg=Single, Mt=Multiple, Avg=(Sm+Md+Lg+Sg+Mt)/5. ZS=Zero-Shot.}
\end{tabular}}
\end{table*}

\section{Experiments}
\label{sec:experiments}

\subsection{Experimental Setup}
\label{sec:exp_setup}

\textbf{Training configuration.}
\modelname{} is trained on 8${\times}$H100 80GB GPUs with DeepSpeed ZeRO-2, bfloat16 mixed precision, AdamW with weight decay 0.05, cosine schedule, and warmup 0.03, batch size 4 per GPU, and maximum sequence length 5,400 tokens. The FGVLA module, MLP connector, and LLM undergo full fine-tuning. Stage~1 trains on LLaVA-OneVision-Data, LLaVA-Instruct-150K, and optical RS data ($\text{lr}=2{\times}10^{-5}$). Stage~2 introduces non-optical modalities with full optical RS replay ($\text{lr}=5{\times}10^{-6}$). Both stages run for 4 epochs.

\textbf{Evaluation benchmarks and comparisons.}
Evaluations span two axes: task-specific benchmarks including classification, captioning, visual grounding, VQA, detection, segmentation, change analysis, and visual prompting, as well as modality-comprehensive benchmarks covering optical, infrared, multispectral, SAR, temporal, video, and cross-sensor fusion. Metrics are task-dependent: accuracy and recall for classification, R-1, R-L, MT, BLEU, and CIDEr for captioning, P@0.25, P@0.5, R-1, R-L, and MT for grounded description, P@0.25, P@0.5, and mIoU for grounding and detection, Acc for VQA, R-1, R-L, MT, SS, and SIoU for visual prompting, oIoU and mIoU for segmentation, and F1, IoU, OA, mF1, and mIoU for change detection.

\subsection{Cross-Task Evaluation}
\label{sec:exp_task}
\begin{table}[!t]
\centering
\caption{Comparison of segmentation methods on EarthReason and RRSIS-D testsets. \best{Bold red} denotes the best result and \second{underlined blue} the second best.}
\label{tab:rrsisd}
\tiny
\setlength{\tabcolsep}{5pt}
\renewcommand{\arraystretch}{0.75}
\begin{tabular}{l|cc|cc}
\toprule
\rowcolor{tableheader}
& \multicolumn{2}{c|}{\textbf{Reasoning Seg.}} & \multicolumn{2}{c}{\textbf{Referring Seg.}} \\
\cmidrule(lr){2-3} \cmidrule(lr){4-5}
\rowcolor{tableheader}
\textbf{Model} & \multicolumn{2}{c|}{\textbf{EarthReason}} & \multicolumn{2}{c}{\textbf{RRSIS-D}} \\
\cmidrule(lr){2-3} \cmidrule(lr){4-5}
\rowcolor{tableheader}
& \textbf{oIoU} & \textbf{mIoU} & \textbf{oIoU} & \textbf{mIoU} \\
\midrule
LISA (7B)           & 61.04 & 60.88 & 27.84 & 26.78 \\
\rowcolor{gray!6}
PixelLM (13B)        & 57.94 & 60.01 & 33.89 & 31.65 \\
PSALM (1.3B)          & 66.61 & 68.30 & --    & --    \\
\rowcolor{gray!6}
GeoGround (7B)+SAM  & --    & --    & 61.10 & 60.50 \\
GeoPixel (7B)       & 53.90 & 52.53 & \best{81.77} & 67.99 \\
\rowcolor{gray!6}
SegEarth-R1 (1.3B)    & \second{68.60} & 70.75 & 67.56 & 66.40 \\
RemoteReasoner (7B)+SAM2 & \best{69.02} & \second{70.96} & 54.29 & 50.97 \\
\rowcolor{gray!6}
RemoteSAM (180M)      & --    & --    & \second{80.04} & \second{71.75} \\
\midrule
\rowcolor{oursrow}
\textbf{\modelname{} (2B)}      & 63.41 & 57.97 & 64.12 & 52.82 \\
\rowcolor{oursrow}
\textbf{\modelname{} (2B) + SAM2} & 62.34 & \best{74.20} & 77.22 & \best{71.91} \\
\bottomrule
\end{tabular}
\end{table}

\begin{table}[!t]
    \centering
    \caption{Change detection comparison on WHU-CDC, LEVIR-CDC, and LEVIR-MCI testsets. \best{Bold red} denotes the best result and \second{underlined blue} the second best.}
    \label{tab:change_detection}
    \setlength{\tabcolsep}{3pt}
    \renewcommand{\arraystretch}{0.75}
    \tiny
    \resizebox{\columnwidth}{!}{%
    \begin{tabular}{l|ccc|ccc|ccc}
    \toprule
    \rowcolor{tableheader}
    & \multicolumn{3}{c|}{\textbf{WHU-CDC}} & \multicolumn{3}{c|}{\textbf{LEVIR-CDC}} & \multicolumn{3}{c}{\textbf{LEVIR-MCI}} \\
    \cmidrule(lr){2-4} \cmidrule(lr){5-7} \cmidrule(lr){8-10}
    \rowcolor{tableheader}
    \textbf{Model} & \textbf{F1} & \textbf{IoU} & \textbf{OA} & \textbf{F1} & \textbf{IoU} & \textbf{OA} & \textbf{mF1} & \textbf{mIoU} & \textbf{OA} \\
    \midrule
    \rowcolor{gray!20}
    \textcolor{gray}{SOTA Specialist} & \textcolor{gray}{\best{88.93}} & \textcolor{gray}{\best{81.54}} & \textcolor{gray}{\best{98.01}} & \textcolor{gray}{\best{92.53}} & \textcolor{gray}{\best{86.83}} & \textcolor{gray}{\best{98.54}} & \textcolor{gray}{\best{92.58}} & \textcolor{gray}{\best{86.54}} & \textcolor{gray}{\best{98.22}} \\
    RSUniVLM (1B) & 48.65 & 32.14 & 93.34 & 54.97 & 37.91 & 94.74 & 72.80 & 63.24 & 94.70 \\
    \midrule
    \rowcolor{oursrow}
    \textbf{\modelname{} (2B)} & \second{70.91} & \second{54.93} & \second{97.12} & \second{73.28} & \second{57.83} & \second{96.34} & \second{85.85} & \second{76.39} & \second{96.64} \\
    \bottomrule
    \end{tabular}}
    \end{table}
    
\begin{table}[!t]
    \centering
    \caption{Visual prompting region captioning and classification on OPT-RSVG testset. Following EarthGPT-X, R-1, R-L, and METEOR computed using \texttt{pycocoevalcap}. SS=Semantic Similarity, SIoU=Semantic IoU. \best{Bold red} denotes the best result and \second{underlined blue} the second best.}
    \label{tab:vp_region_cls}
    \scriptsize
    \setlength{\tabcolsep}{4pt}
    \renewcommand{\arraystretch}{0.75}
    \resizebox{\columnwidth}{!}{%
    \begin{tabular}{l|ccc|cc}
    \toprule
    \rowcolor{tableheader}
    & \multicolumn{5}{c}{\textbf{OPT-RSVG}} \\
    \cmidrule(lr){2-6}
    \rowcolor{tableheader}
    \textbf{Model} & \multicolumn{3}{c|}{\textbf{Region Captioning}} & \multicolumn{2}{c}{\textbf{Region Cls.}} \\
    \cmidrule(lr){2-4} \cmidrule(lr){5-6}
    \rowcolor{tableheader}
    & \textbf{R-1} & \textbf{R-L} & \textbf{MT} & \textbf{SS} & \textbf{SIoU} \\
    \midrule
    InternVL3 (8B)  & 14.71 & 14.28 & 18.64 & 76.89 & 66.61 \\
    \rowcolor{gray!6}
    Sphinx (13B)        & 31.19 & 30.28 & 12.95 & 84.32 & 77.18 \\
    EarthGPT (13B)      & 32.47 & 31.52 & 15.29 & 87.45 & 81.23 \\
    \midrule
    Sphinx-V (13B)      & 33.04 & 32.08 & 14.10 & 80.71 & 72.17 \\
    \rowcolor{gray!6}
    VIP-LLaVA (7B)     & 22.15 & 21.36 & 8.47  & 68.24 & 58.91 \\
    EarthGPT-X (13B)    & 68.88 & 66.87 & 39.61 & 97.28 & 96.32 \\
    \midrule
    \rowcolor{oursrow}
    \textbf{\modelname{} (2B) (box)} & \second{73.94} & \second{72.66} & \second{45.18} & \second{97.76} & \second{97.32} \\
    \rowcolor{oursrow}
    \textbf{\modelname{} (2B) (VP)}  & \best{75.71} & \best{74.55} & \best{47.11} & \best{98.40} & \best{98.42} \\
    \bottomrule
    \end{tabular}}
    \end{table}

    \begin{figure*}[!t]
        \centering
        \includegraphics[width=\textwidth, height=0.75\textheight]{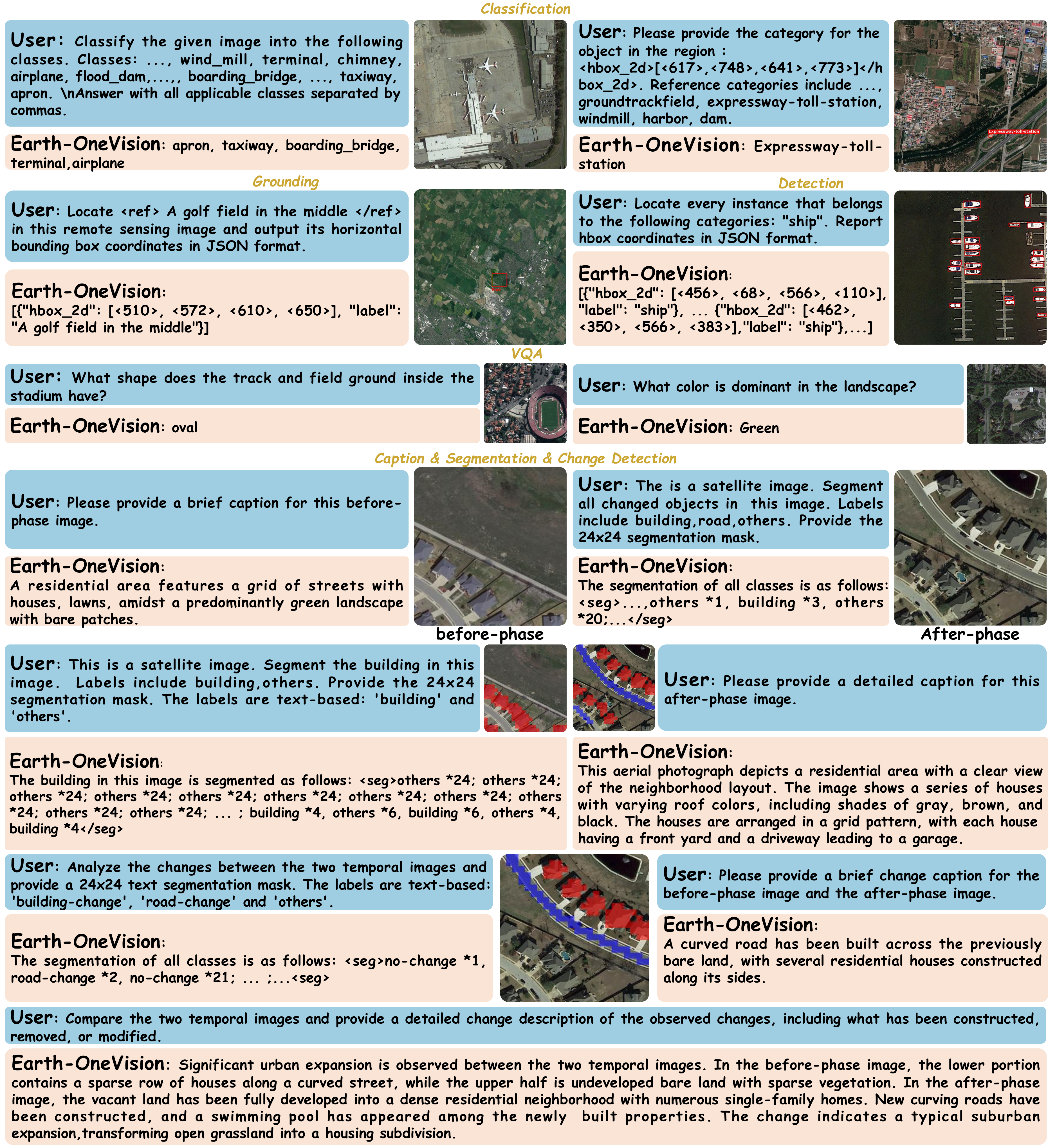}
        \caption{Cross-task inference visualization on optical imagery, covering multi-label classification, region classification, grounding, object detection, VQA, captioning, segmentation, change detection, and change captioning.}
        \label{fig:vis_task}
        \end{figure*}

This subsection evaluates \modelname{} across eight task categories to verify that a single unified model handles coordinate regression, discrete classification, free-form text generation, and pixel-level mask prediction.

\subsubsection{Classification}

Classification tasks span image-level scene recognition, multi-label land-cover classification, and multi-temporal disaster assessment, as shown in Table~\ref{tab:classification}. \modelname{} achieves 97.31\% accuracy on WHU-RS19~\cite{balestra2025whu} testset and 78.03\% recall on BigEarthNet RGB~\cite{helber2018introducing}, surpassing EarthDial by 1.10\% and 5.00\%. On xBD~\cite{irvin2024teochat} testset, \modelname{} achieves an average accuracy of 80.89\%, outperforming EarthDial by 6.89\%. On FMoW~\cite{christie2018functional} temporal classification testset, \modelname{} achieves 73.63\% accuracy, outperforming EarthDial by 3.60\%, demonstrating consistent gains across scene, multi-label, and temporal classification settings.

\subsubsection{Captioning}

Captioning tasks cover change captioning, image captioning, temporal captioning, and region captioning, as shown in Table~\ref{tab:captioning_comprehensive}. \modelname{} surpasses EarthDial on LEVIR-MCI~\cite{liu2024change} testset with 76.45\% METEOR vs.\ 74.80\%, on MUDS testset with 37.40\% METEOR vs.\ 33.56\% and 33.98\% ROUGE-1 vs.\ 28.16\%, on xBD temporal captioning testset with 92.33\% ROUGE-1 vs.\ 87.26\%, and on SAR-Ship region captioning testset with 78.63\% ROUGE-1 vs.\ 63.10\%, showing consistent captioning improvements across change, image, temporal, and region subtasks.

\subsubsection{Visual Grounding}

Visual grounding evaluates precise localization from textual descriptions across five benchmarks, as shown in Table~\ref{tab:visual_grounding}. \modelname{} achieves 94.41\% P@0.5 on DIOR-RSVG~\cite{zhan2023rsvg} testset, surpassing the SOTA specialist at 84.35\%, 87.52\% P@0.5 on OPT-RSVG~\cite{li2024language} testset, surpassing the SOTA specialist at 86.51\%, 90.77\% P@0.5 on VRSBench-VG~\cite{li2024vrsbench} testset, and 82.36\% P@0.5 on RSVG-HR~\cite{lan2024language} testset. On the RSVG testset, \modelname{} achieves 65.06\% P@0.5, falling below the SOTA specialist score of 84.48\% due to extremely small targets, yet it achieves SOTA on three of five benchmarks, demonstrating strong spatial localization from language descriptions.

\subsubsection{Visual Question Answering}

Visual question answering (VQA) covers diverse question types such as category, existence, position, quantity, scene, and color across three benchmarks, as shown in Table~\ref{tab:rsvqa}. \modelname{} achieves 80.32\% accuracy on VRSBench-VQA~\cite{li2024vrsbench} testset, outperforming RSThinker by 4.42\%, 86.36\% accuracy on RSVQA-HR~\cite{lobry2020rsvqa} testset, outperforming EarthDial by 15.36\%, and 92.91\% accuracy on RSVQA-LR~\cite{lobry2020rsvqa} testset, showing strong question-answering performance across diverse question types.

\subsubsection{Object Detection}

Object detection evaluates simultaneous localization and category recognition across varying scales, as shown in Table~\ref{tab:earthdial_comprehensive}. \modelname{} achieves 40.63\% P@0.5 on GeoChat-Instruct~\cite{kuckreja2023geochat} testset, outperforming EarthDial by 14.64\%, 59.23\% P@0.5 on NWPU-VHR-10~\cite{cheng2017remote} testset, outperforming EarthDial by 42.58\%, 20.02\% P@0.5 on SwimmingPool testset, and 19.73\% P@0.5 on UrbanTreeCrown testset, demonstrating robust generalization across object densities.

\subsubsection{Segmentation}

\modelname{} is evaluated on EarthReason~\cite{li2025segearth} for reasoning segmentation and RRSIS-D~\cite{liu2024rotated} for referring segmentation testsets, as shown in Table~\ref{tab:rrsisd}. R-RLE tokens achieve 63.41\% oIoU and 57.97\% mIoU on EarthReason, and 64.12\% oIoU and 52.82\% mIoU on RRSIS-D. SAM2~\cite{ravi2024sam} refinement improves the EarthReason score to 74.20\% mIoU, outperforming RemoteReasoner by 3.24\%, and the RRSIS-D score to 71.91\% mIoU, outperforming RemoteSAM by 0.16\%, enabling competitive segmentation without a dedicated mask decoder.

\subsubsection{Change Detection}

Change detection evaluates pixel-level difference localization between multi-temporal image pairs, as shown in Table~\ref{tab:change_detection}. \modelname{} achieves 70.91\% F1 on WHU-CDC~\cite{shi2024whucdc} testset, 73.28\% F1 on LEVIR-CDC testset, and 85.85\% mF1 on LEVIR-MCI testset, outperforming RSUniVLM by 22.26\%, 18.31\%, and 13.05\%, though a gap remains versus specialist models due to R-RLE boundary resolution limits.

\subsubsection{Visual Prompting}

Visual prompting (VP) evaluates region-level understanding when targets are specified via visual cues, as shown in Table~\ref{tab:vp_region_cls}. \modelname{} (VP) achieves 75.71\% ROUGE-1 on OPT-RSVG region captioning testset, outperforming EarthGPT-X by 6.83\%, as well as 98.40\% SS and 98.42\% SIoU on region classification, surpassing EarthGPT-X by 1.12\% and 2.10\%, showing that visual-cue-specified region understanding surpasses prior RS-MLLMs.

Figure~\ref{fig:vis_task} shows a single unified model handling classification, grounding, object detection, VQA, captioning, segmentation, and change detection on optical imagery within the same autoregressive framework. Across all eight task categories, \modelname{} consistently matches or surpasses 4B-72B RS-MLLMs, demonstrating that unified multi-task training yields complementary rather than competing gains.

\subsection{Cross Sensor-Modality Evaluation}
\label{sec:exp_modality}

\modelname{} is assessed along six sensor dimensions to examine performance consistency across different imaging physics: optical on VHM-HnstD~\cite{pang2025vhm} and FIT-RSRC~\cite{luo2024skysensegpt}, infrared on HIT-UAV~\cite{suo2023hit}, multispectral on BigEarthNet-MS, SoSAT-LCZ42, and TreeSatAI~\cite{soni2025earthdial}, SAR on SARLANG-Bench~\cite{wei2026sarlang}, temporal on DVL-Bench~\cite{xuan2025dynamicvl}, video on ERA~\cite{mou2020era} and CapERA~\cite{bashmal2023capera}, and cross-sensor fusion on EarthMind-Bench~\cite{earthmind2025}.

\subsubsection{Optical Understanding}

To evaluate fine-grained optical understanding, \modelname{} is assessed on VHM-HnstD and FIT-RSRC benchmarks, as shown in Table~\ref{tab:fitrs}. On VHM-HnstD, \modelname{} (2B) achieves 86.86\% accuracy, surpassing VHM (7B) by 4.92\%. On FIT-RSRC, average accuracy reaches 70.89\%, outperforming SkySenseGPT (7B) by 15.44\%, demonstrating that the 2B model achieves superior fine-grained optical understanding over larger counterparts.

\begin{table*}[!t]
\centering
\caption{Optical evaluation on VHM-HnstD and FIT-RSRC benchmarks. \best{Bold red} denotes the best result and \second{underlined blue} the second best.}
\label{tab:fitrs}
\scriptsize
\resizebox{\textwidth}{!}{
\begin{tabular}{l|ccccccccc|ccccc}
\toprule
\rowcolor{tableheader}
& \multicolumn{9}{c|}{\textbf{VHM-HnstD}} & \multicolumn{5}{c}{\textbf{FIT-RSRC}} \\
\cmidrule(lr){2-10} \cmidrule(lr){11-15}
\rowcolor{tableheader}
\textbf{Model} & \textbf{Pres.-Fact.} & \textbf{Col-Fact.} & \textbf{Col-Dec.-Ex.} & \textbf{Col-Dec.-Pan.} & \textbf{AbsPos-Fact.} & \textbf{AbsPos-Dec.} & \textbf{RelPos-Fact.} & \textbf{RelPos-Dec.} & \cellcolor{avgcol}\textbf{Avg} & \textbf{Subj.} & \textbf{Obj.} & \textbf{Rel.} & \textbf{Exist.} & \cellcolor{avgcol}\textbf{Avg} \\
\midrule
LLaVA-1.5 (7B) & 70.40 & 66.96 & 23.33 & 42.00 & 61.61 & 12.00 & 34.71 & 31.67 & \cellcolor{avgcol}42.84 & 6.4 & 11.2 & 21.0 & 7.8 & \cellcolor{avgcol}11.60 \\
\rowcolor{gray!6}
CogVLM (7B) & 74.71 & 31.25 & \second{68.00} & \best{100.00} & 33.93 & 16.67 & 29.34 & 11.00 & \cellcolor{avgcol}45.61 & -- & -- & -- & -- & \cellcolor{avgcol}-- \\
Qwen-VL-Chat (7B) & 72.99 & 47.62 & 8.33 & 39.00 & 54.29 & 29.52 & 31.79 & 28.91 & \cellcolor{avgcol}39.06 & -- & -- & -- & -- & \cellcolor{avgcol}-- \\
\rowcolor{gray!6}
TinyLLaVA (3B) & -- & -- & -- & -- & -- & -- & -- & -- & \cellcolor{avgcol}-- & 22.0 & 18.0 & 12.0 & 55.6 & \cellcolor{avgcol}26.90 \\
LLaVA-HR (7B) & -- & -- & -- & -- & -- & -- & -- & -- & \cellcolor{avgcol}-- & 7.8 & 13.4 & 32.6 & 47.8 & \cellcolor{avgcol}25.40 \\
\rowcolor{gray!6}
GeoChat (7B) & -- & -- & -- & -- & -- & -- & -- & -- & \cellcolor{avgcol}-- & 7.6 & 8.6 & 25.6 & 45.6 & \cellcolor{avgcol}21.85 \\
VHM (7B) & \second{85.06} & \second{81.50} & \best{93.33} & \second{93.00} & \second{76.79} & \best{90.67} & \second{47.52} & \best{87.67} & \cellcolor{avgcol}\second{81.94} & -- & -- & -- & -- & \cellcolor{avgcol}-- \\
\rowcolor{gray!6}
SkySenseGPT (7B) & -- & -- & -- & -- & -- & -- & -- & -- & \cellcolor{avgcol}-- & \second{25.2} & \second{54.0} & \second{50.4} & \second{92.2} & \cellcolor{avgcol}\second{55.45} \\
\midrule
\rowcolor{oursrow}
\textbf{\modelname{} (2B)} & \best{87.07} & \best{85.75} & \best{93.33} & \best{100.00} & \best{83.93} & \second{90.00} & \best{73.14} & \second{81.67} & \cellcolor{avgcol}\best{86.86} & \best{67.16} & \best{62.98} & \best{58.23} & \best{95.20} & \cellcolor{avgcol}\best{70.89} \\
\bottomrule
\end{tabular}}
\end{table*}
\begin{table}[ht]
    \centering
    \caption{SAR evaluation on SARLANG-Bench. VQA uses GPT-4o-mini semantic matching. BLEU, METEOR, ROUGE-L, and CIDEr computed using \texttt{pycocoevalcap}. \best{Bold red} denotes the best result and \second{underlined blue} the second best.}
    \label{tab:sarlang}
    \scriptsize
    \setlength{\tabcolsep}{3pt}
    \renewcommand{\arraystretch}{0.75}
    \resizebox{\columnwidth}{!}{%
    \begin{tabular}{l|cccc|cccc|c}
    \toprule
    \rowcolor{tableheader}
    & \multicolumn{4}{c|}{\textbf{Captioning (Complex)}} & \multicolumn{4}{c|}{\textbf{Captioning (Concise)}} & \textbf{VQA} \\
    \cmidrule(lr){2-5} \cmidrule(lr){6-9} \cmidrule(lr){10-10}
    \rowcolor{tableheader}
    \textbf{Model} & \textbf{B-1} & \textbf{B-4} & \textbf{R-L} & \textbf{CIDEr} & \textbf{B-1} & \textbf{B-4} & \textbf{R-L} & \textbf{CIDEr} & \textbf{Acc} \\
    \midrule
    LLaVA-1.5 (13B) & 34.90 & 12.01 & 32.43 & 45.13 & 25.42 & 7.20 & 22.65 & 11.03 & 70.04 \\
    \rowcolor{gray!6}
    LLaVA-1.5 (7B) & 35.24 & 12.70 & 32.72 & 46.35 & 23.97 & 5.71 & 21.52 & 10.27 & 70.30 \\
    Qwen2-VL (7B) & \second{35.78} & \second{13.08} & \second{32.84} & 48.36 & 24.58 & 7.11 & 22.28 & 9.56 & 68.55 \\
    \rowcolor{gray!6}
    Qwen2.5-VL (7B) & 32.79 & 12.25 & 30.24 & \second{55.64} & \second{37.75} & \second{22.36} & \second{38.50} & \second{70.05} & \second{73.33} \\
    InternVL2.5 (8B) & 28.11 & 7.51 & 29.66 & 16.15 & 29.50 & 9.85 & 30.28 & 4.84 & 39.90 \\
    \midrule
    \rowcolor{oursrow}
    \textbf{\modelname{} (2B)} & \best{46.26} & \best{20.31} & \best{42.37} & \best{110.24} & \best{43.06} & \best{22.96} & \best{41.20} & \best{91.83} & \best{80.68} \\
    \bottomrule
    \end{tabular}}
    \end{table}
    \begin{table}[ht]
        \centering
        \caption{Infrared evaluation on HIT-UAV testset. Following EarthDial, R-1, R-L, and METEOR computed using NLTK. \best{Bold red} and \second{underlined blue} denote the best and second-best results.}
        \label{tab:hit_uav_infrared}
        \scriptsize
        \setlength{\tabcolsep}{3.5pt}
        \renewcommand{\arraystretch}{0.75}
        \resizebox{\columnwidth}{!}{%
        \begin{tabular}{l|ccccc|ccc}
        \toprule
        \rowcolor{tableheader}
        & \multicolumn{5}{c|}{\textbf{Grounded Description}} & \multicolumn{3}{c}{\textbf{Region Captioning}} \\
        \cmidrule(lr){2-6} \cmidrule(lr){7-9}
        \rowcolor{tableheader}
        \textbf{Model} & \textbf{P@0.25} & \textbf{P@0.5} & \textbf{R-1} & \textbf{R-L} & \textbf{MT} & \textbf{R-1} & \textbf{R-L} & \textbf{MT} \\
        \midrule
        GPT-4o       & 0.7   & 0.1   & 14.20 & 10.56 & 7.16  & 10.96 & 9.02  & 8.23  \\
        \rowcolor{gray!6}
        InternVL2 (4B) & 6.4   & 0.6   & 28.1  & 27.68 & \second{23.94} & 11    & 9.53  & 8.40  \\
        GeoChat (7B)      & 8.0   & 0.8   & 22.82 & 22.22 & 22.27 & 59.85 & 59.85 & 51.31 \\
        \rowcolor{gray!6}
        EarthDial (4B)    & \second{13.86} & \second{2.61} & \second{28.31} & \second{28.06} & 22.25 & \second{61.83} & \second{61.83} & \second{52.80} \\
        \midrule
        \rowcolor{oursrow}
        \textbf{\modelname{} (2B)} & \best{79.48} & \best{22.75} & \best{68.29} & \best{64.18} & \best{55.92} & \best{91.62} & \best{91.61} & \best{90.24} \\
        \bottomrule
        \end{tabular}}
        \end{table}
        \begin{table}[!t]
            \centering
            \caption{Multispectral evaluation on classification benchmarks (\%). \best{Bold red} denotes the best result and \second{underlined blue} the second best.}
            \label{tab:multispectral_cls}
            \scriptsize
            \setlength{\tabcolsep}{4pt}
            \renewcommand{\arraystretch}{0.75}
            \resizebox{\columnwidth}{!}{%
            \begin{tabular}{l|ccc}
            \toprule
            \rowcolor{tableheader}
             & \textbf{BigEarthNet-MS} & \textbf{SoSAT-LCZ42} & \textbf{TreeSatAI} \\
            \cmidrule(lr){2-2}\cmidrule(lr){3-3}\cmidrule(lr){4-4}
            \rowcolor{tableheader}
            \textbf{Model} & \textbf{Recall} & \textbf{Acc} & \textbf{Acc} \\
            \midrule
            GPT-4o          & 49 & 15.53 & 16.73 \\
            \rowcolor{gray!6}
            EarthMind (4B)       & \second{71.20} & 59.20 & -- \\
            EarthDial (4B)       & 69.94 & \best{60.72} & \best{56.61} \\
            \midrule
            \rowcolor{oursrow}
            \textbf{\modelname{} (2B)} & \best{75.74} & \second{59.44} & \second{56.17} \\
            \bottomrule
            \end{tabular}}
            \end{table}
            \begin{table}[!t]
                \centering
                \caption{Temporal analysis evaluation on DVL-Bench. BCA for Basic Change Analysis, CSE for Change Speed Estimation: single-choice (SQ) and multi-choice (MQ) accuracy (\%). \best{Bold red} denotes the best result and \second{underlined blue} the second best.}
                \label{tab:dvlbench}
                \scriptsize
                \renewcommand{\arraystretch}{0.50}
                \resizebox{\columnwidth}{!}{%
                \begin{tabular}{l|cccc|c}
                \toprule
                \rowcolor{tableheader}
                \textbf{Model} & \textbf{BCA-SQ} & \textbf{BCA-MQ} & \textbf{CSE-SQ} & \textbf{CSE-MQ} & \cellcolor{avgcol}\textbf{Avg} \\
                \midrule
                o4-mini & 62.8 & 36.1 & 33.8 & 12.4 & \cellcolor{avgcol}\second{36.28} \\
                GPT-4.1 & \second{66.1} & \second{39.7} & 31.3 & 5.4 & \cellcolor{avgcol}35.63 \\
                GPT-4o & 63.3 & 19.3 & 32.3 & 7.3 & \cellcolor{avgcol}30.55 \\
                Gemini-2.5-Flash & 46.3 & 15.8 & 21.0 & 12.1 & \cellcolor{avgcol}23.80 \\
                \midrule
                TEOChat (7B) & 35.1 & 8.7 & 17.0 & 10.8 & \cellcolor{avgcol}17.90 \\
                EarthDial (4B) & 62.2 & 20.3 & 30.9 & 12.2 & \cellcolor{avgcol}31.40 \\
                LLaVA-OV (72B) & 59.9 & 6.5 & 25.9 & 6.2 & \cellcolor{avgcol}24.63 \\
                InternVL3 (14B) & 63.2 & 15.3 & 28.8 & 4.0 & \cellcolor{avgcol}27.83 \\
                Qwen2.5-VL (32B) & 62.0 & 33.3 & \second{36.9} & 3.2 & \cellcolor{avgcol}33.85 \\
                Qwen2.5-VL (72B) & 65.4 & 24.3 & 34.6 & 4.0 & \cellcolor{avgcol}32.08 \\
                DVLChat (7B) & 64.9 & 21.3 & 31.3 & \best{18.6} & \cellcolor{avgcol}34.03 \\
                \midrule
                \rowcolor{oursrow}
                \textbf{\modelname{} (2B)} & \best{68.61} & \best{50.63} & \best{37.80} & \second{12.83} & \cellcolor{avgcol}\best{42.47} \\
                \bottomrule
                \end{tabular}}
                \end{table}
                \begin{table}[!t]
                    \centering
                    \caption{Video understanding evaluation on ERA and CapERA testsets. ERA classification: Accuracy (\%). CapERA captioning: BLEU, METEOR, ROUGE-L, and CIDEr via \texttt{pycocoevalcap}. \best{Bold red} denotes the best result and \second{underlined blue} the second best.}
                    \label{tab:era}
                    \scriptsize
                    \setlength{\tabcolsep}{4pt}
                    \renewcommand{\arraystretch}{0.75}
                    \resizebox{\columnwidth}{!}{%
                    \begin{tabular}{l|c|ccccc}
                    \toprule
                    \rowcolor{tableheader}
                    & \textbf{Video Classification} & \multicolumn{5}{c}{\textbf{Video Captioning}} \\
                    \cmidrule(lr){2-2}\cmidrule(lr){3-7}
                    \rowcolor{tableheader}
                    \textbf{Model} & \textbf{ERA} & \multicolumn{5}{c}{\textbf{CapERA}} \\
                    \cmidrule(lr){2-2}\cmidrule(lr){3-7}
                    \rowcolor{tableheader}
                    & \textbf{Acc} & \textbf{B-1} & \textbf{B-4} & \textbf{MT} & \textbf{R-L} & \textbf{CIDEr} \\
                    \midrule
                    \rowcolor{gray!30}
                    \textcolor{gray}{SOTA Specialist} & \textcolor{gray}{84.70} & \textcolor{gray}{\best{50.43}} & \textcolor{gray}{\best{22.90}} & \textcolor{gray}{--} & \textcolor{gray}{\best{43.90}} & \textcolor{gray}{\second{60.42}} \\
                    \midrule
                    MiniGPT-v2 (7B)        & 60.34 & 28.7 & 4.3 & 14.7 & 25.8 & 15.4 \\
                    \rowcolor{gray!6}
                    SkyEyeGPT (7B)         & 55.54 & 30.8 & 7.9 & \second{16.9} & 28.6 & 27.4 \\
                    MiniCPM-V-2.6 (8B)     & 78.80 & 35.0 & 6.2 & 14.5 & 28.8 & 29.6 \\
                    \rowcolor{gray!6}
                    Gemini-2.5-Flash  & \second{84.91} & 38.8 & 10.2 & 16.8 & 32.3 & 32.9 \\
                    LLaVA-1.5-UAV (7B)     & 76.52 & 43.0 & 14.7 & 15.9 & 37.5 & 54.9 \\
                    \midrule
                    \rowcolor{oursrow}
                    \textbf{\modelname{} (2B)} & \best{89.14} & \second{44.32} & \second{17.16} & \best{19.94} & \second{39.33} & \best{71.73} \\
                    \bottomrule
                    \end{tabular}}
                    \end{table}
                    \begin{table*}[!t]
                        \centering
                        \caption{Multi-modality evaluation on EarthMind-Bench. MCQ for multiple-choice question tasks including scene classification, object existence, hallucination detection, object counting, and spatial relationship, scored out of 100. OE for open-ended tasks including image caption, disaster forecasting, route planning, and urban assessment, GPT-4 scored (1-5). M-Avg and O-Avg: MCQ and OE averages. $^\dagger$Closed-source. \best{Bold red} denotes the best result and \second{underlined blue} the second best.}
                        \label{tab:earthmind_oe}
                        \scriptsize
                        \renewcommand{\arraystretch}{0.82}
                        \setlength{\tabcolsep}{5pt}
                        \resizebox{\textwidth}{!}{%
                        \begin{tabular}{l ccccc c cccc c}
                        \toprule
                        \rowcolor{tableheader}
                        & \multicolumn{5}{c}{\textbf{MCQ Tasks}} & & \multicolumn{4}{c}{\textbf{OE Tasks}} & \\
                        \cmidrule(lr){2-6} \cmidrule(lr){7-7} \cmidrule(lr){8-11} \cmidrule(lr){12-12}
                        \rowcolor{tableheader}
                        \textbf{Model} & \makecell{\textbf{Scene}\\\textbf{Class.}} & \makecell{\textbf{Object}\\\textbf{Exist.}} & \makecell{\textbf{Halluci.}\\\textbf{Detect.}} & \makecell{\textbf{Object}\\\textbf{Count.}} & \makecell{\textbf{Spatial}\\\textbf{Relation.}} & \cellcolor{avgcol}\textbf{M-Avg} & \makecell{\textbf{Image}\\\textbf{Captioning}} & \makecell{\textbf{Disaster}\\\textbf{Forecast.}} & \makecell{\textbf{Route}\\\textbf{Plann.}} & \makecell{\textbf{Urban}\\\textbf{Assess.}} & \cellcolor{avgcol}\textbf{O-Avg} \\
                        \midrule
                        \multicolumn{12}{l}{\cellcolor{gray!12}Evaluation on Optical} \\
                        GPT-4o$^\dagger$ & \second{67.8} & \second{79.9} & \best{86.4} & 34.0 & \second{52.3} & \cellcolor{avgcol}\second{64.1} & \best{4.58} & 1.75 & \second{2.01} & 2.18 & \cellcolor{avgcol}2.63 \\
                        \rowcolor{gray!6}
                        GeoChat (7B) & 40.9 & 51.8 & 46.8 & 18.9 & 19.0 & \cellcolor{avgcol}35.5 & 1.92 & 1.73 & 1.33 & 2.14 & \cellcolor{avgcol}1.78 \\
                        GeoPixel (7B) & 55.3 & 67.8 & 73.6 & 33.5 & 34.7 & \cellcolor{avgcol}53.0 & 2.80 & 1.80 & 1.68 & 1.95 & \cellcolor{avgcol}2.06 \\
                        \rowcolor{gray!6}
                        EarthDial (4B) & 58.2 & 72.4 & 75.9 & 40.6 & 44.2 & \cellcolor{avgcol}58.3 & 2.82 & 1.95 & 1.66 & 2.10 & \cellcolor{avgcol}2.13 \\
                        EarthMind (4B) & 64.3 & 77.5 & 83.6 & \second{50.1} & 33.1 & \cellcolor{avgcol}61.7 & 3.35 & \second{3.37} & \second{2.01} & \second{2.55} & \cellcolor{avgcol}\second{2.82} \\
                        \rowcolor{oursrow}
                        \textbf{\modelname{} (2B)} & \best{96.24} & \best{81.88} & \second{83.89} & \best{54.31} & \best{87.18} & \cellcolor{avgcol}\best{80.70} & \second{3.99} & \best{3.89} & \best{2.65} & \best{2.66} & \cellcolor{avgcol}\best{3.30} \\
                        \midrule
                        \multicolumn{12}{l}{\cellcolor{gray!12}Evaluation on SAR} \\
                        GPT-4o$^\dagger$ & 35.2 & 71.4 & 72.9 & 22.8 & 36.6 & \cellcolor{avgcol}47.8 & 2.89 & 3.05 & 1.65 & 2.04 & \cellcolor{avgcol}2.40 \\
                        \rowcolor{gray!6}
                        GeoChat (7B) & 28.6 & 49.8 & 46.8 & 27.9 & 18.5 & \cellcolor{avgcol}34.3 & 1.78 & 1.65 & 1.25 & 1.56 & \cellcolor{avgcol}1.56 \\
                        GeoPixel (7B) & 35.2 & 59.0 & 65.7 & 30.5 & 32.8 & \cellcolor{avgcol}44.6 & 2.08 & 1.97 & 1.45 & 1.68 & \cellcolor{avgcol}1.80 \\
                        \rowcolor{gray!6}
                        EarthDial (4B) & 40.6 & 65.3 & 69.2 & 35.7 & 36.4 & \cellcolor{avgcol}49.4 & 2.26 & 2.09 & 1.60 & 1.86 & \cellcolor{avgcol}1.95 \\
                        EarthMind (4B) & \second{64.4} & \best{77.4} & \second{74.6} & \second{46.8} & \second{43.1} & \cellcolor{avgcol}\second{61.3} & \second{3.10} & \second{3.25} & \second{1.89} & \second{2.30} & \cellcolor{avgcol}\second{2.64} \\
                        \rowcolor{oursrow}
                        \textbf{\modelname{} (2B)} & \best{91.94} & \second{73.15} & \best{75.83} & \best{53.67} & \best{85.90} & \cellcolor{avgcol}\best{76.10} & \best{3.52} & \best{3.41} & \best{2.51} & \best{2.83} & \cellcolor{avgcol}\best{3.07} \\
                        \midrule
                        \multicolumn{12}{l}{\cellcolor{gray!12}Evaluation on Optical-SAR Fusion} \\
                        GPT-4o$^\dagger$ & 64.8 & 79.6 & \second{86.2} & 31.6 & 43.5 & \cellcolor{avgcol}61.1 & 3.68 & 1.59 & 1.82 & 2.03 & \cellcolor{avgcol}2.28 \\
                        \rowcolor{gray!6}
                        GPT-4V$^\dagger$ & 30.2 & 64.8 & 62.4 & 32.8 & 39.8 & \cellcolor{avgcol}46.0 & 2.89 & 1.48 & 1.57 & 1.79 & \cellcolor{avgcol}1.93 \\
                        EarthMind (4B) & \second{65.5} & \second{84.4} & \best{88.1} & \second{52.4} & \second{59.7} & \cellcolor{avgcol}\second{70.0} & \second{3.80} & \second{3.37} & \second{2.21} & \second{2.70} & \cellcolor{avgcol}\second{3.02} \\
                        \rowcolor{oursrow}
                        \textbf{\modelname{} (2B)} & \best{96.24} & \best{85.23} & 85.78 & \best{53.99} & \best{88.46} & \cellcolor{avgcol}\best{81.94} & \best{3.99} & \best{4.01} & \best{2.64} & \best{3.10} & \cellcolor{avgcol}\best{3.44} \\
                        \bottomrule
                        \end{tabular}}
                        \end{table*}

                        \begin{figure*}[!t]
                            \centering
                            \includegraphics[width=\textwidth, height=0.4\textheight]{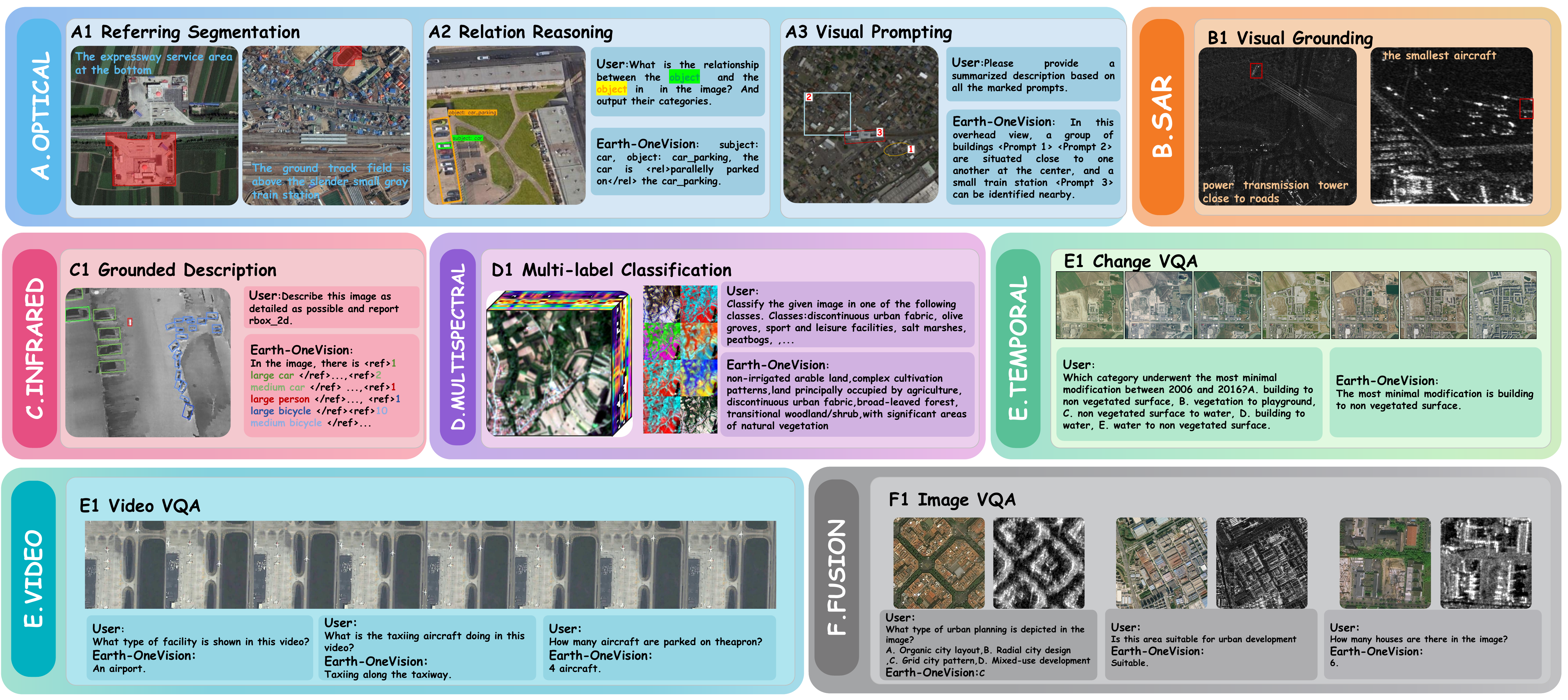}
                            \caption{Cross sensor-modality inference visualization across six sensor modalities and cross-sensor fusion. (A) Optical: referring segmentation, relation reasoning, and visual prompting. (B) SAR: visual grounding. (C) Infrared: grounded description. (D) Multispectral: multi-label classification. (E) Temporal: change VQA. (F) Video: video VQA. (G) Cross-sensor fusion: image VQA.}
                            \label{fig:vis_modality}
                        \end{figure*}
\subsubsection{SAR Interpretation}

To assess SAR modality comprehension, \modelname{} is evaluated on SARLANG-Bench~\cite{wei2026sarlang}, as shown in Table~\ref{tab:sarlang}. \modelname{} achieves 80.68\% accuracy on VQA, 110.24 CIDEr on complex captioning, and 91.83 CIDEr on concise captioning, leading all three SAR subtasks with 3.5$\times$ fewer parameters than the next-best model.

\subsubsection{Infrared Perception}

Infrared imagery captures thermal radiation signatures invisible to optical sensors. \modelname{} is evaluated on HIT-UAV testset, as shown in Table~\ref{tab:hit_uav_infrared}, covering grounded description and region captioning tasks. On HIT-UAV, \modelname{} achieves 79.48\% P@0.25 on grounded description, surpassing EarthDial by 65.62\%, and 91.62\% ROUGE-1 on region captioning, surpassing EarthDial by 29.79\%, indicating robust generalization to thermal infrared.

\subsubsection{Multispectral Recognition}

Multispectral bands are grouped into pseudo-RGB images and fed as multi-image input, as shown in Table~\ref{tab:multispectral_cls}. \modelname{} achieves 75.74\% recall on BigEarthNet-MS testset, outperforming EarthDial by 5.80\%, 59.44\% accuracy on SoSAT-LCZ42 testset, and 56.17\% accuracy on TreeSatAI testset, showing competitive multispectral generalization without modality-specific components.

\subsubsection{Temporal Analysis}

For temporal reasoning and change analysis, \modelname{} is evaluated on DVL-Bench, as shown in Table~\ref{tab:dvlbench}, which covers multi-temporal change QA across BCA and CSE tasks in single-choice and multi-choice formats. \modelname{} achieves the best overall average of 42.47\%, surpassing o4-mini by 6.19\% and DVLChat by 8.44\%, with 68.61\% accuracy on BCA single-choice and 50.63\% accuracy on BCA multi-choice leading all models, demonstrating strong temporal reasoning under both formats.

\subsubsection{Video Understanding}

Table~\ref{tab:era} reports video understanding results on ERA~\cite{mou2020era} and CapERA~\cite{bashmal2023capera} testsets, UAV video datasets with 25 event categories. \modelname{} achieves 89.14\% video classification accuracy on ERA, surpassing both the SOTA specialist at 84.70\% and Gemini-2.5-Flash at 84.91\%, and 71.73 CIDEr on CapERA, outperforming all compared methods without a dedicated video encoder.

\subsubsection{Cross Sensor-Modality Reasoning}

To thoroughly examine cross sensor-modality reasoning, \modelname{} is evaluated on EarthMind-Bench~\cite{earthmind2025}, as shown in Table~\ref{tab:earthmind_oe}, covering optical, SAR, and fusion modalities with multiple-choice questions (MCQ) and open-ended QA. \modelname{} achieves 80.70\% M-Avg on optical, 76.10\% M-Avg on SAR, and 81.94\% M-Avg on cross-sensor fusion, surpassing EarthMind by 19.00\%, 14.80\%, and 11.94\%, with fusion achieving the highest MCQ average, demonstrating that cross-sensor fusion consistently outperforms single-modality inputs.

Figure~\ref{fig:vis_modality} presents representative outputs across six sensor modalities and cross-sensor fusion, covering object detection, segmentation, grounding, VQA, captioning, and change detection on optical, SAR, infrared, multispectral, temporal, video, and cross-sensor fusion inputs, validating the unified design across fundamentally different imaging physics.

\begin{table*}[t]
\centering
\caption{Ablation studies on six representative benchmarks. Cross-task benchmarks evaluate spatial and reasoning capabilities across tasks. Cross sensor-modality benchmarks evaluate generalization across sensor modalities and tasks. MS: multispectral. \best{Bold red} denotes the best result and \second{underlined blue} the second best.}
\label{tab:ablation}
\scriptsize
\renewcommand{\arraystretch}{0.75}
\setlength{\tabcolsep}{5pt}
\resizebox{\textwidth}{!}{%
\begin{tabular}{ccc|ccc|ccc}
\toprule
\rowcolor{tableheader}
& & & \multicolumn{3}{c|}{\textbf{\textit{Cross-Task}}} & \multicolumn{3}{c}{\textbf{\textit{Cross Sensor-Modality}}} \\
\rowcolor{tableheader}
\textbf{FGVLA} & \textbf{SLIS} & \textbf{PCMA} &
\makecell{\textbf{OPT-RSVG}\\\textbf{Grounding}\\\textbf{(P@0.5)}} &
\makecell{\textbf{WHU-RS19}\\\textbf{Classification}\\\textbf{(Acc)}} &
\makecell{\textbf{RSVQA-HR}\\\textbf{VQA}\\\textbf{(Acc)}} &
\makecell{\textbf{SARLANG}\\\textbf{SAR $\cdot$ VQA}\\\textbf{(Acc)}} &
\makecell{\textbf{CapERA}\\\textbf{Video $\cdot$ Captioning}\\\textbf{(METEOR)}} &
\makecell{\textbf{SoSAT-LCZ42}\\\textbf{MS $\cdot$ Classification}\\\textbf{(Acc)}} \\
\midrule
$\times$ & \checkmark & \checkmark & 84.73 & 95.42 & 84.38 & \second{80.09} & \best{20.83} & 57.83 \\
\rowcolor{gray!6}
\checkmark & $\times$ & \checkmark & 85.34 & \best{97.71} & \second{85.67} & 77.47 & 18.63 & \best{60.17} \\
\checkmark & \checkmark & $\times$ & \second{85.91} & 95.82 & 84.12 & 76.86 & 17.48 & 58.91 \\
\midrule
\rowcolor{oursrow}
\checkmark & \checkmark & \checkmark & \best{87.52} & \second{97.31} & \best{86.36} & \best{80.68} & \second{19.94} & \second{59.44} \\
\bottomrule
\end{tabular}}
\end{table*}

\subsection{Ablation Studies}
\label{sec:exp_ablation}

To verify the necessity of each core design choice, Table~\ref{tab:ablation} evaluates three components by selectively disabling each one in turn, across six representative benchmarks spanning cross-task and cross sensor-modality axes.

FGVLA aligns multi-level visual features with the language model through cross-scale attention and early-layer feature injection. Removing it drops OPT-RSVG P@0.5 by 2.79\% and RSVQA-HR average accuracy by 1.98\%.

SLIS represents spatial positions as dedicated coordinate tokens. Removing it drops OPT-RSVG grounding P@0.5 by 2.18\%, while classification remains largely unaffected, indicating that dedicated coordinate tokens primarily benefit spatial reasoning.

PCMA consolidates optical understanding before progressively incorporating non-optical modalities with optical replay. Removing it causes the largest drops, with SARLANG-Bench VQA accuracy dropping by 3.82\% and CapERA METEOR dropping by 2.46\%.

The three ablation results confirm that FGVLA, PCMA, and SLIS each address a distinct bottleneck, with the full model achieving the best overall results across both task and modality axes.

\section{Conclusion}

This paper presents \modelname{}, a unified RS-MLLM that brings broad-spectrum earth observation interpretation under a single 2B-parameter autoregressive framework, spanning six sensor modalities and cross-sensor fusion across 9 task categories. Unlike existing RS-MLLMs, \modelname{} introduces FGVLA, SLIS, and PCMA to jointly resolve all three core bottlenecks within a single framework. For joint training across all modalities, \datasetname{} covers over 34M QA pairs across six sensor modalities and cross-sensor fusion across 9 task categories, surpassing prior RS instruction datasets in both scale and modality breadth. Evaluated across cross-task and cross sensor-modality benchmarks, \modelname{} matches or surpasses larger RS-MLLMs across tasks, demonstrating that a single unified model can handle the full spectrum of RS tasks and sensor modalities. The core finding is that cross-modality synergy among diverse physical sensors consistently drives performance gains in multi-task and multi-sensor-modality RS interpretation. Current limitations include the spatial resolution ceiling of text-based mask encoding and incomplete coverage of hyperspectral and LiDAR modalities. Future work will pursue reasoning-enhanced scaling and broader modality extension.

\bibliographystyle{IEEEtran}
\bibliography{references}

\end{document}